\documentclass[]{bytedance_seed}

\usepackage[toc,page,header]{appendix}

\usepackage{minitoc}
\usepackage{amsthm}
\newtheorem{theorem}{Theorem}
\usepackage[table]{xcolor}

\usepackage{amsmath,amsfonts,bm}

\def\Figref#1{Figure~\ref{#1}}

\def\eqref#1{equation~\ref{#1}}

\def\1{\bm{1}}

\def\rr{{\textnormal{r}}}

\def\ru{{\textnormal{u}}}

\def\rvx{{\mathbf{x}}}

\def\vx{{\bm{x}}}

\DeclareMathAlphabet{\mathsfit}{\encodingdefault}{\sfdefault}{m}{sl}
\SetMathAlphabet{\mathsfit}{bold}{\encodingdefault}{\sfdefault}{bx}{n}

\newcommand{\E}{\mathbb{E}}

\usepackage{titletoc}

\usepackage{float}

\title{Do Visual RL Gains Scale at Test Time? A Quantile View of Post-Training}

\author[1]{Austin Wang}
\author[1]{Lexing Ying}

\affiliation[1]{ByteDance Seed}

\abstract{
Reinforcement learning (RL) post-training improves the single-sample quality of visual generative models, but deployment often provides additional inference compute through Best-of-$N$ sampling. Whether RL gains transfer under such test-time scaling, particularly on difficult held-out tasks, remains poorly understood in the visual domain. \textbf{Our central contribution is to systematically identify and characterize a failure mode of scaling transfer in visual RL:} across multiple RL methods, model backbones, and evaluations spanning compositional generation, text rendering, and preference alignment, higher single-sample reward does not necessarily imply a better quality-compute frontier. RL gains may attenuate, saturate, or even reverse at large sampling budgets, with the base model eventually outperforming its post-trained counterpart. We explain these regimes through a quantile-transfer framework, showing that Best-of-$N$ places increasing weight on the upper tail of the reward distribution as $N$ grows. Consequently, RL can improve typical samples while compressing or degrading the rare high-quality outcomes that test-time scaling exploits. As a simple intervention motivated by this analysis, we introduce Quantile Utility Policy Optimization (QUPO), which optimizes a controllable mixture of quantile utilities corresponding to different inference budgets. Experiments on SD3.5-M and BAGEL show that QUPO better preserves upper-tail quality and substantially improves scaling on challenging held-out prompts relative to standard RL post-training. Our results suggest that visual generative models should be evaluated and optimized not only for single-sample quality, but for how post-training reshapes the full reward distribution that inference-time scaling exploits.
}

\correspondence{Austin Wang at \email{austinwang@bytedance.com}}

\checkdata[Code and Model Checkpoints]{\url{https://github.com/austin-k-wang/QUPO}}

\begin{document}
\maketitle

\section{Introduction}

Reinforcement learning (RL) post-training has become an effective way to improve visual generative models, including diffusion and flow-based generators~\cite{ho2020denoising,karras2022elucidating,albergo2025stochastic,lipman2022flow}, and unified multimodal models such as BAGEL~\cite{deng2025emerging}. Recent methods improve compositional accuracy, text rendering, and preference alignment~\cite{liu2026flow,zheng2026diffusionnft,xue2025advantage}, but are typically optimized and evaluated primarily through single-sample reward. This leaves open how visual post-training affects a model's ability to benefit from additional inference-time compute.

At deployment, additional compute can be converted into quality through Best-of-$N$ sampling: generate $N$ candidates and select the highest-reward output. We ask: \textbf{when RL improves Best-of-$1$, do those gains persist as $N$ increases?} We call this property \textit{scaling transfer}, the extent to which post-training gains are preserved along the Best-of-$N$ quality--compute frontier. This question is especially important on difficult held-out prompts, where additional sampling is most useful.

We find that scaling transfer is far from guaranteed. Across multiple visual RL methods and tasks, Best-of-$1$ gains can persist, attenuate, or even reverse as $N$ grows. On held-out compositional benchmarks such as GenEval2~\cite{kamath2025geneval} and ConceptMix~\cite{wu2024conceptmix}, several post-trained models outperform the base model at small $N$ but are eventually overtaken at larger budgets. Text rendering instead exhibits strongly attenuated but positive transfer, while preference-alignment gains are substantially more persistent. These qualitatively different regimes are obscured by single-sample evaluation.

We explain these behaviors through a quantile view of the per-prompt reward distribution. Let $Q_{\pi}(u\mid c)$ denote the reward quantile function of model $\pi$ on prompt $c$. Our analysis shows that Best-of-$1$ weights reward quantiles uniformly, whereas increasing $N$ places progressively greater emphasis on the upper tail. Post-training can therefore improve typical generations while leaving rare high-reward outcomes unchanged or degraded, causing its advantage to shrink or reverse at larger sampling budgets. Empirically, the observed quantile shifts closely match the corresponding scaling-transfer regimes.

Motivated by this analysis, we introduce \textbf{Quantile Utility Policy Optimization} (QUPO) as a simple intervention for the scaling-transfer failure identified in visual RL. QUPO is a drop-in objective that optimizes a controllable mixture of Best-of-$k$ utilities, explicitly emphasizing the reward quantiles that additional inference samples exploit. Related language model work has explored Pass@$k$ and other tail-aware objectives~\cite{walder2026pass,tajwar2026maximum,ramasubramanian2026tail}; our main contribution centers on systematically identifying and characterizing scaling-transfer failure in visual generation, with QUPO demonstrating that a simple change to the training utility can mitigate it. Across SD3.5-M and BAGEL, QUPO better preserves upper-tail reward quality and substantially improves Best-of-$N$ scaling relative to standard RL post-training.

We summarize our contributions as follows:
\begin{itemize}
    \item We identify and characterize \textit{scaling transfer} in visual RL, showing that post-training gains can persist, attenuate, or reverse as test-time sampling compute increases.
    
    \item We develop a quantile-transfer framework showing that larger Best-of-$N$ budgets increasingly emphasize the upper reward tail, and empirically connect upper-tail changes to observed scaling behavior.
    
    \item We introduce QUPO, a simple quantile-aware post-training objective that improves scaling transfer on both SD3.5-M and BAGEL.
\end{itemize}
\section{Related Work}

\subsection{Reward-Based Post-Training of Visual Generators}

Diffusion and flow-based generators have been post-trained through direct reward backpropagation~\citep{prabhudesai2023aligning,clark2024directly}, reinforcement learning~\citep{black2024training,fan2023dpok,xue2025dancegrpo,liu2026flow,zheng2026diffusionnft,choi2026rethinking,xue2025advantage,wang2026aligning}, and preference optimization~\citep{wallace2024diffusion,wang2026beyond,lu2025inpo,lu2025smoothed,han2026discrete}. These methods primarily target single-sample alignment through expected-reward or preference objectives and are typically evaluated at fixed inference budgets. Consequently, how post-training reshapes generation quality across larger sampling budgets remains poorly understood.

Reward-based post-training has also been extended to unified multimodal models (UMMs), which jointly support language, visual understanding, and image generation. Recently, UniGRPO \cite{liu2026unigrpo} applies GRPO-style optimization to models such as BAGEL~\cite{deng2025emerging}. We study both flow-based generators and UMMs to test whether scaling-transfer behavior persists across distinct generative architectures.

\subsection{Test-Time Scaling of Visual Generators}

Best-of-$N$ (BoN) inference improves generation quality by sampling multiple candidates and selecting the highest-scoring output, providing a simple characterization of the quality--compute frontier. More sophisticated approaches use reward-gradient guidance, search and lookahead, or posterior and reward-tilted sampling~\citep{singhal2025general,uehara2025inference,huang2026guide,bansal2024universal,yeh2025training,zhang2025improving,zheng2025blade,wang2025ensemble,zheng2024ensemble}. Whereas these methods study how to better use inference compute for a fixed model, we study how post-training changes the model's inherent ability to benefit from that compute. We therefore use Best-of-$N$ as a model-agnostic diagnostic of the generator's native reward distribution, avoiding confounding from a particular search or guidance procedure.

\subsection{Test-Time-Aware Policy Optimization}

Recent language model works study how post-training affects multi-sample inference: PKPO~\cite{walder2026pass} optimizes Pass@$k$, MaxRL~\cite{tajwar2026maximum} introduces compute-indexed objectives, MaxPO~\cite{takashiro2026advantage} improves Max@$k$ estimators, and TEA~\cite{li2026should} extrapolates upper-tail reward statistics to deployment budgets exceeding the rollout budget. \cite{cheng2026rl} analyzes how RL reshapes LLM reasoning, linking changes in strategy selection to cross-domain transfer, reasoning coverage, and scaling laws in RL compute. Concurrently, TailRL~\cite{ramasubramanian2026tail} develops a tail-sensitive objective expressible as a harmonic mixture of Best-of-$k$ utilities. QUPO subsumes this objective as a special case while allowing controllable weighting across inference budgets. In visual generation, Multi-Axis Max@$K$~\cite{onoda2026multi} targets representative diversity through a set-level objective that maximizes scores across samples separately for predefined semantic modes, such as color and lighting.

Our primary contribution is the systematic characterization of \textit{scaling transfer} in visual RL: when Best-of-$1$ gains persist or fail as inference budgets grow. We connect these failures to reward-quantile shifts and derive QUPO from this quantile view of the scalar Best-of-$N$ quality--compute frontier, demonstrating a simple remedy across visual generation architectures.

\section{Background}

\subsection{Reward Distributions and Best-of-$N$}

Let $\pi(\vx\mid c)$ denote a visual generative model conditioned on prompt $c$, with scalar reward $r(\vx,c)$. Sampling $\rvx\sim\pi(\cdot\mid c)$ induces a reward random variable $\rr=r(\rvx,c)$ with CDF
\[
F_\pi(t\mid c)=\mathbb{P}(\rr\leq t),
\]
and quantile function
\begin{equation}
Q_\pi(u\mid c)
=
\inf\{t:F_\pi(t\mid c)\geq u\},
\qquad u\in(0,1).
\end{equation}

For $N$ independent generations
$\rvx_1,\ldots,\rvx_N\sim\pi(\cdot\mid c)$, define the expected Best-of-$N$ reward
\begin{equation}
B_N^\pi
=
\E_c
\E_{\rvx_1,\ldots,\rvx_N\sim\pi(\cdot\mid c)}
\left[
\max_i r(\rvx_i,c)
\right].
\end{equation}
For $N=1$,
\begin{equation}
B_1^\pi
=
\E_c\E_{\rvx\sim\pi(\cdot\mid c)}[r(\rvx,c)],
\end{equation}
so expected single-sample reward is the first point on the Best-of-$N$ quality--compute frontier.

Given a reference model $\pi_{\mathrm{ref}}$ and post-trained model $\pi_{\mathrm{RL}}$, we define
\begin{equation}
\label{eq:bon_adv}
A_N
=
B_N^{\pi_{\mathrm{RL}}}
-
B_N^{\pi_{\mathrm{ref}}}.
\end{equation}
A decreasing $A_N$ indicates attenuating scaling transfer, while $A_N<0$ indicates frontier reversal. In Sec.~\ref{sec:quantile_analysis}, we relate $A_N$ to changes in the reward quantiles.

\subsection{Reward-Maximizing Visual Post-Training}

Many visual RL methods optimize learning signals derived from rewards assigned to individual generations. The canonical reward objective is
\begin{equation}
\mathcal{J}_{\mathrm{reward}}(\theta)
=
\E_c\E_{\rvx\sim\pi_\theta(\cdot\mid c)}
[r(\rvx,c)]
=
B_1^{\pi_\theta}.
\end{equation}
Several methods sample groups of $G$ generations and construct normalized advantages such as
\begin{equation}
\label{eq:grpo_adv}
\widehat{A}_i
=
\frac{r_i-\bar r}{s_r+\epsilon},
\qquad
r_i=r(\rvx_i,c),
\end{equation}
where $\bar r$ and $s_r$ are the group mean and standard deviation. FlowGRPO~\cite{liu2026flow}, GRPO-Guard~\cite{wang2026grpo}, AWM~\cite{xue2025advantage}, and UniGRPO~\cite{liu2026unigrpo} use this advantage in GRPO-style~\cite{shao2024deepseekmath} optimization, while DiffusionNFT~\cite{zheng2026diffusionnft} uses a different update rule but similarly derives its signal from per-generation rewards.

Crucially, using multiple rollouts during training does not itself optimize a Best-of-$G$ utility: the group provides normalization or credit assignment, while utility remains defined per sample. These methods are therefore primarily aligned with single-sample reward rather than explicitly optimizing the multi-sample utility realized at deployment.

\section{Diagnosing Scaling Transfer in Visual RL} \label{sec:diagnostic}

We study whether gains from visual RL post-training persist as test-time sampling compute increases on difficult held-out prompts. Across compositional generation, text rendering, and preference alignment, we compare each post-trained SD3.5-M model against its matched base model using up to $N=128$ samples per prompt.

\subsection{Experimental Protocol}
\label{sec:diagnostic_protocol}

We evaluate FlowGRPO~\cite{liu2026flow}, DiffusionNFT~\cite{zheng2026diffusionnft}, GRPO-Guard~\cite{wang2026grpo}, and AWM~\cite{xue2025advantage}, where applicable. Models are either reproduced using the official implementation or obtained from officially released LoRA adapters (Table~\ref{tab:diagnostic_settings}). Each model is post-trained on a source task and evaluated on harder held-out prompts from the same capability domain.

For each prompt $c$, we sample $G=128$ generations and estimate Best-of-$N$ performance by averaging the maximum reward over all size-$N$ subsets:
\begin{equation}
\widehat{B}_N(c)
=
\frac{1}{\binom{G}{N}}
\sum_{\substack{S\subseteq\{1,\ldots,G\}\\|S|=N}}
\max_{i\in S} r_i .
\end{equation}
This U-statistic is unbiased for expected Best-of-$N$ reward and uses all available samples. We use \textit{oracle Best-of-$N$}, selecting candidates according to the evaluation reward itself. Importantly, all evaluation rewards are reference-free and can also be used for practical inference-time reranking without ground-truth image annotations. Thus, our protocol characterizes both the intrinsic reward distribution and the attainable Best-of-$N$ frontier under the corresponding reward-based selector, excluding scoring overhead. We average $\widehat{B}_N(c)$ across prompts and compare each post-trained model against its matched base model. Extended experimental details are provided in Section~\ref{app:diagnostic-details}.

\subsection{Compositional Generation}
\label{sec:diagnostic_compositional}

We evaluate models post-trained on GenEval~\cite{ghosh2023geneval} on the harder held-out GenEval2 benchmark~\cite{kamath2025geneval}. DiffusionNFT and AWM use a CFG scale of $1.0$, while FlowGRPO and GRPO-Guard use CFG $4.5$; we report the base model under both settings.

\begin{figure*}[t]
    \centering
    \begin{subfigure}[t]{0.48\linewidth}
        \centering
        \includegraphics[width=\linewidth]
            {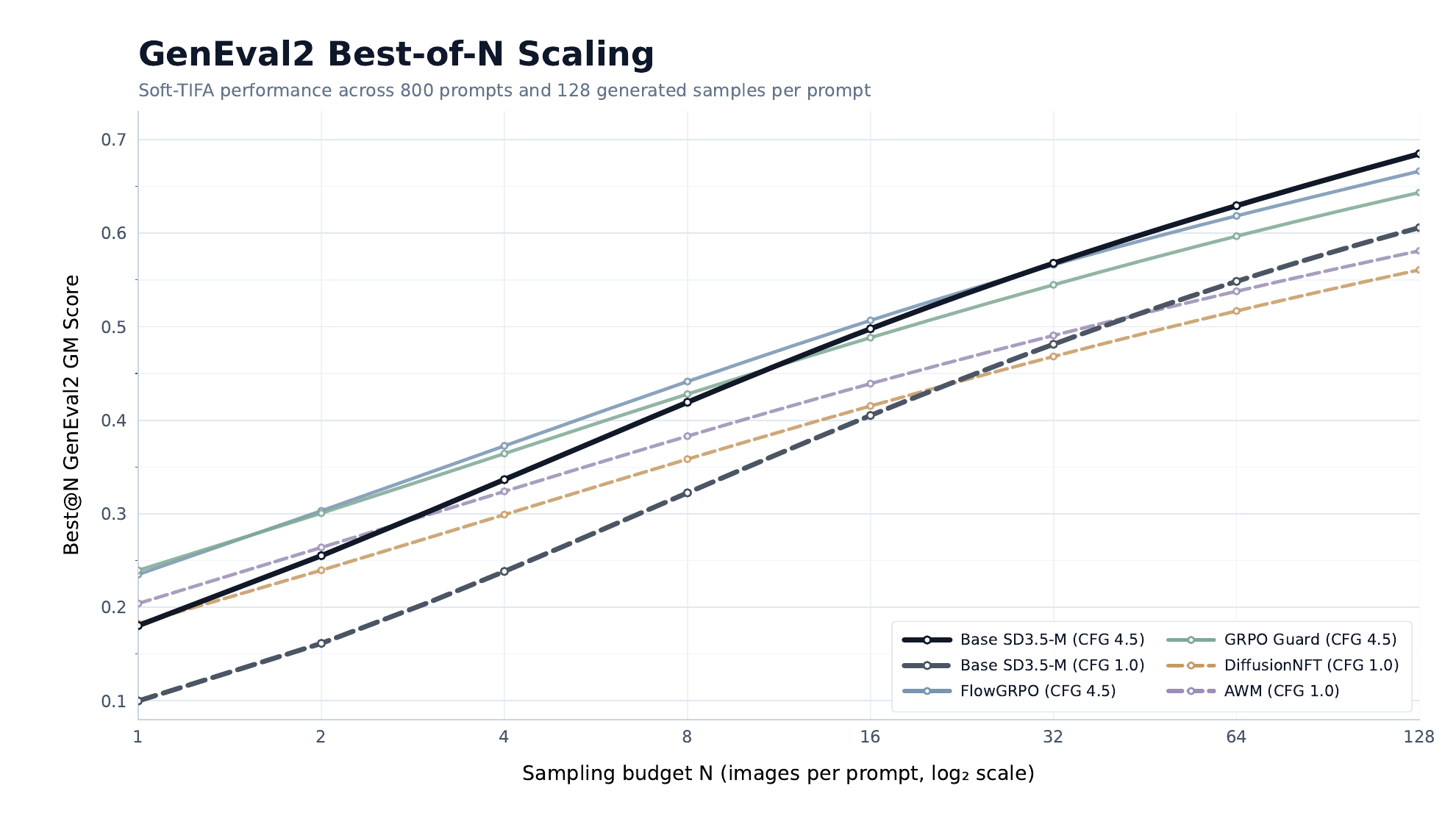}
        \caption{GenEval2.}
        \label{fig:geneval2_bon}
    \end{subfigure}
    \hfill
    \begin{subfigure}[t]{0.48\linewidth}
        \centering
        \includegraphics[width=\linewidth]
            {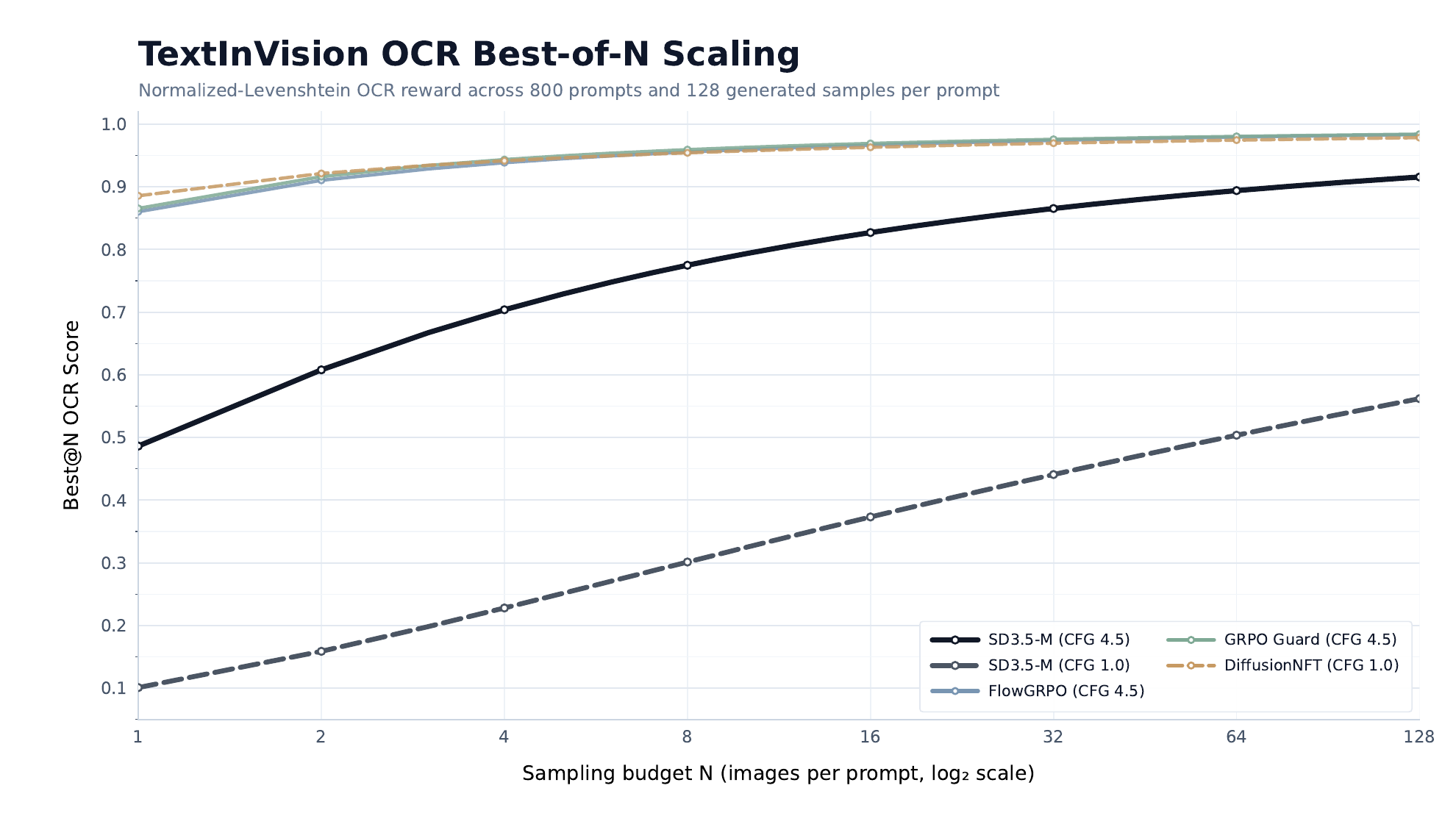}
        \caption{TextInVision.}
        \label{fig:textinvision_bon}
    \end{subfigure}

    \par\medskip

    \begin{subfigure}[t]{0.48\linewidth}
        \centering
        \includegraphics[width=\linewidth]
            {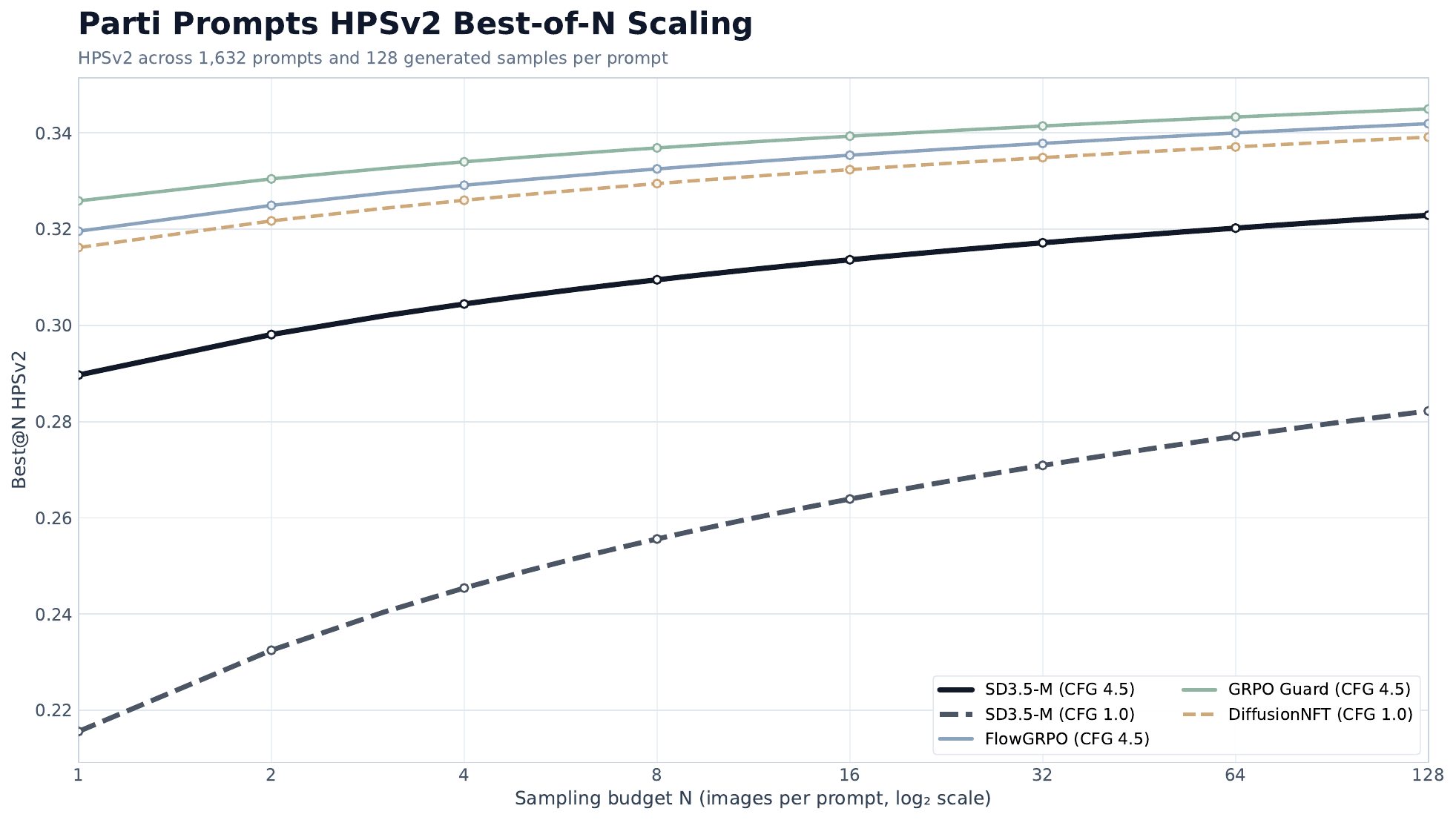}
        \caption{Parti-Prompts.}
        \label{fig:parti_bon}
    \end{subfigure}

    \caption{\textbf{Best-of-$N$ scaling on difficult held-out prompts.}
    \textbf{(a)} Compositional generation exhibits frontier reversal.
    \textbf{(b)} Text-rendering gains remain positive but decay toward
    saturation.
    \textbf{(c)} Preference-alignment gains remain approximately persistent
    as $N$ increases.}
    \label{fig:diagnostic_bon}
\end{figure*}

\Figref{fig:diagnostic_bon}a shows a clear frontier reversal. All post-trained models improve Best-of-$1$ and small-$N$ performance, but their advantage steadily decreases and eventually becomes negative at large $N$. Thus, successful compositional post-training does not necessarily preserve the high-reward outcomes exploited by larger sampling budgets.



\subsection{Text Rendering}
\label{sec:diagnostic_ocr}

We evaluate models trained on the FlowGRPO OCR task on harder held-out TextInVision prompts~\cite{fallah2025textinvision}. Unlike compositional generation, post-trained models remain uniformly stronger, but their advantage decreases substantially as they approach the reward ceiling and the base model catches up under additional sampling. We refer to this regime as \textit{positive but decaying scaling transfer}.

\subsection{Preference Alignment}
\label{sec:diagnostic_preference}

We train on Pick-a-Pic prompts with PickScore~\cite{kirstain2023pick} and evaluate on held-out Parti Prompts using HPSv2~\cite{wu2023human}, testing transfer across both prompts and reward models. Here, post-trained models remain uniformly stronger and maintain a similar advantage as $N$ increases, exhibiting \textit{approximately persistent scaling transfer}.

\subsection{Scaling-Transfer Regimes}
\label{sec:scaling_regimes}

Across the three domains, scaling transfer ranges from \textit{frontier reversal} to \textit{positive but decaying} and \textit{approximately persistent} transfer. These results indicate that weakened Best-of-$N$ scaling is neither universal nor explained by single-sample performance alone. Instead, the shape of the entire reward distribution must determine how post-training gains evolve with inference compute. In the next section, we develop a quantile-based analysis that makes this connection explicit.

\section{Understanding Scaling Transfer Through Reward Quantiles}
\label{sec:quantile_analysis}

\subsection{Best-of-$N$ as Upper-Tail Quantile Transfer}
\label{sec:quantile_transfer}

To explain the scaling-transfer regimes above, we study how post-training reshapes the per-prompt reward distribution. Define the average {quantile lift}
\begin{equation}
\Delta Q(u)
=
\E_c\!\left[
Q_{\pi_{\mathrm{RL}}}(u\mid c)
-
Q_{\pi_{\mathrm{ref}}}(u\mid c)
\right],
\qquad u\in(0,1).
\label{eq:quantile_lift}
\end{equation}

\begin{theorem}[Best-of-$N$ as quantile transfer]
\label{thm:bon_quantile}
Assuming integrable rewards, for every $N\geq1$, the post-training advantage at inference budget $N$ (Eq.\ref{eq:bon_adv}) can be expressed as
\begin{equation}
A_N
=
\int_0^1 \Delta Q(u)\,Nu^{N-1}\,du
=
\E_{\ru_N\sim\mathrm{Beta}(N,1)}
[\Delta Q(\ru_N)].
\label{eq:an_quantile}
\end{equation}
\end{theorem}

The proof is deferred to Appendix~\ref{app:proof_theorem1}. For $N=1$, reward quantiles are weighted uniformly; as $N$ increases, $\mathrm{Beta}(N,1)$ concentrates near $u=1$, placing increasing emphasis on the upper tail. Thus, post-training can improve average reward while degrading large-$N$ performance if its gains do not extend to the highest reward quantiles.

\begin{figure}[t]
    \centering
    \begin{subfigure}[t]{0.32\linewidth}
        \centering
        \includegraphics[width=\linewidth]{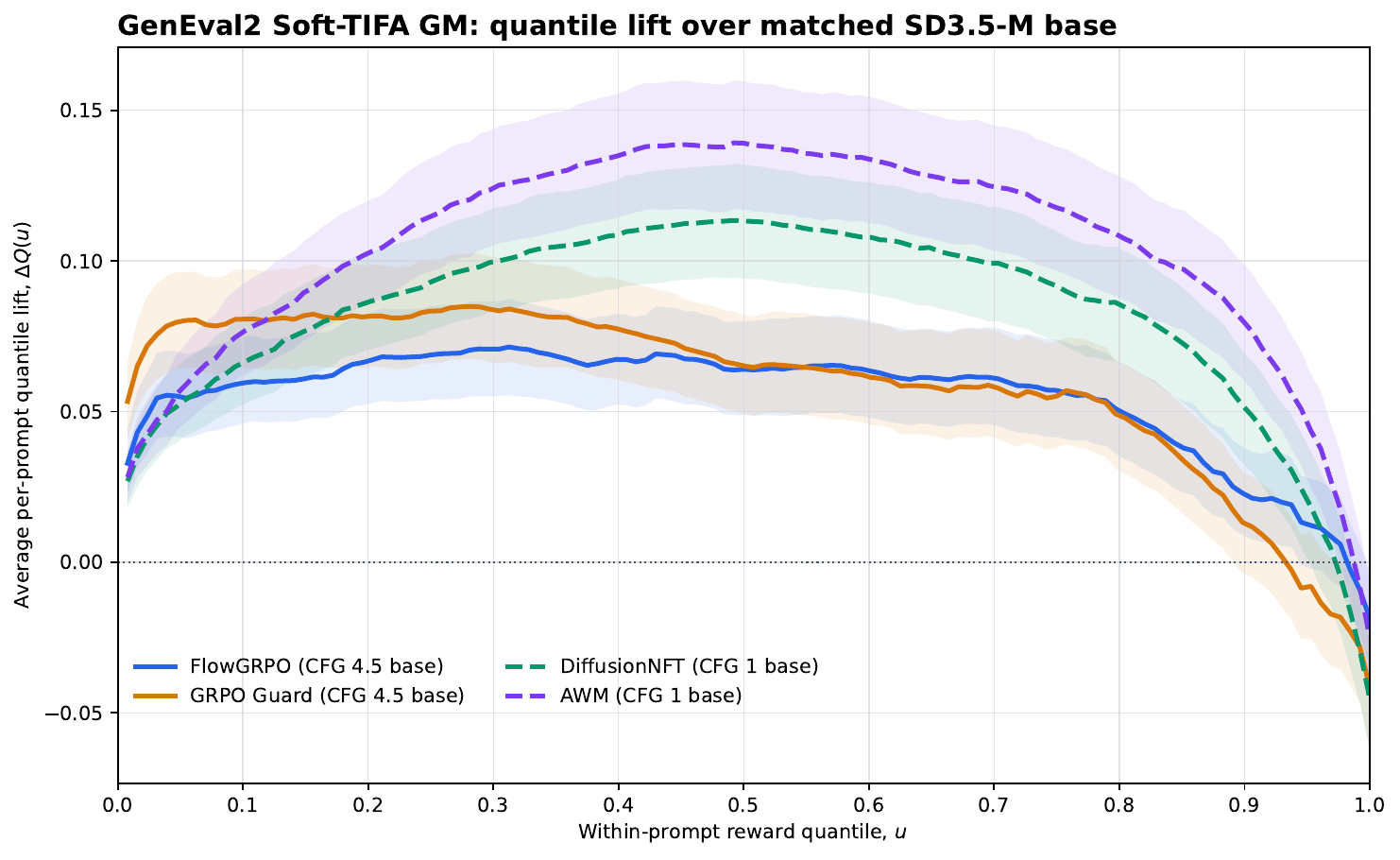}
        \caption{GenEval2.}
        \label{fig:geneval2_quantile_lift}
    \end{subfigure}
    \hfill
    \begin{subfigure}[t]{0.32\linewidth}
        \centering
        \includegraphics[width=\linewidth]{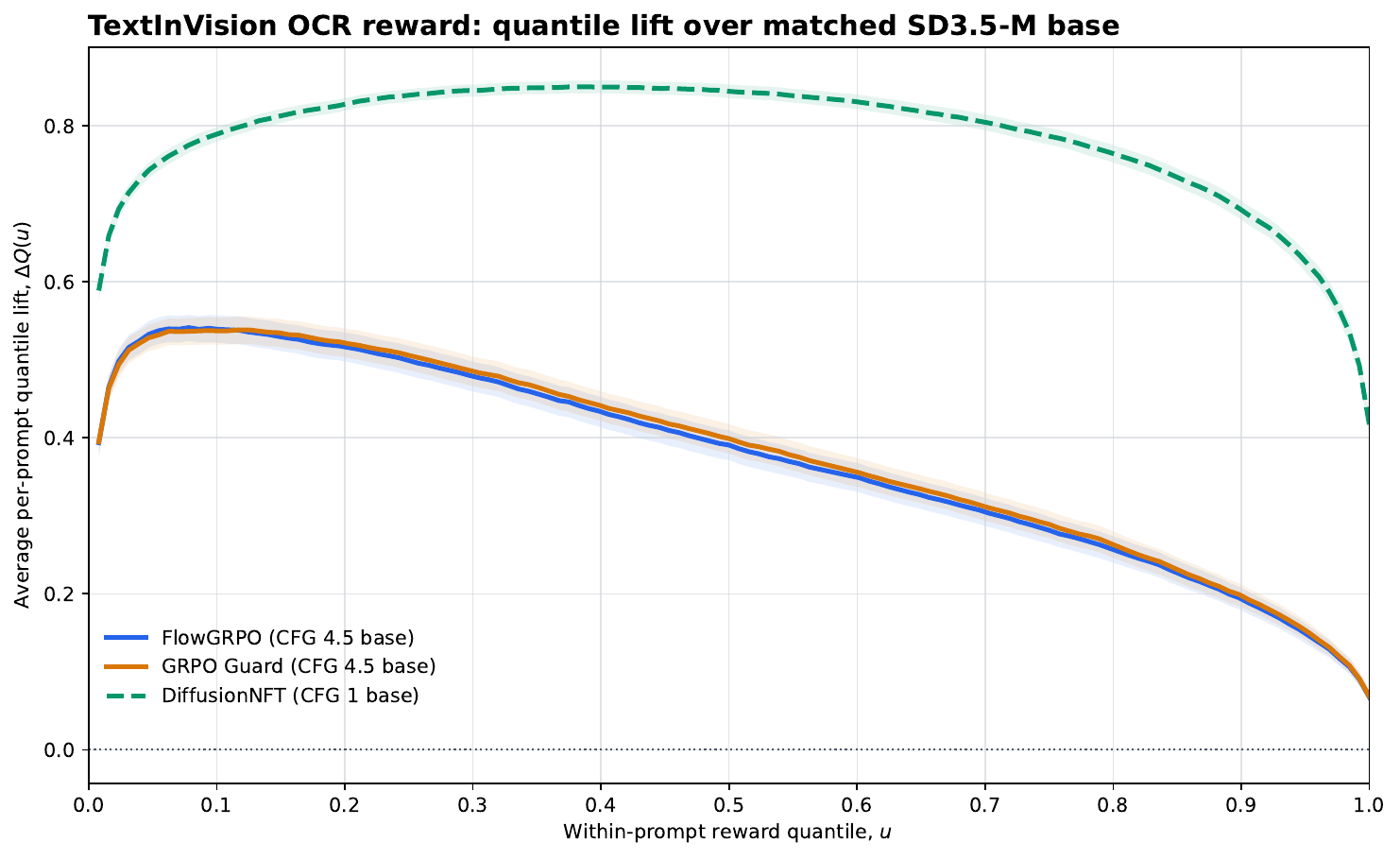}
        \caption{TextInVision.}
        \label{fig:textinvision_quantile_lift}
    \end{subfigure}
    \hfill
    \begin{subfigure}[t]{0.32\linewidth}
        \centering
        \includegraphics[width=\linewidth]{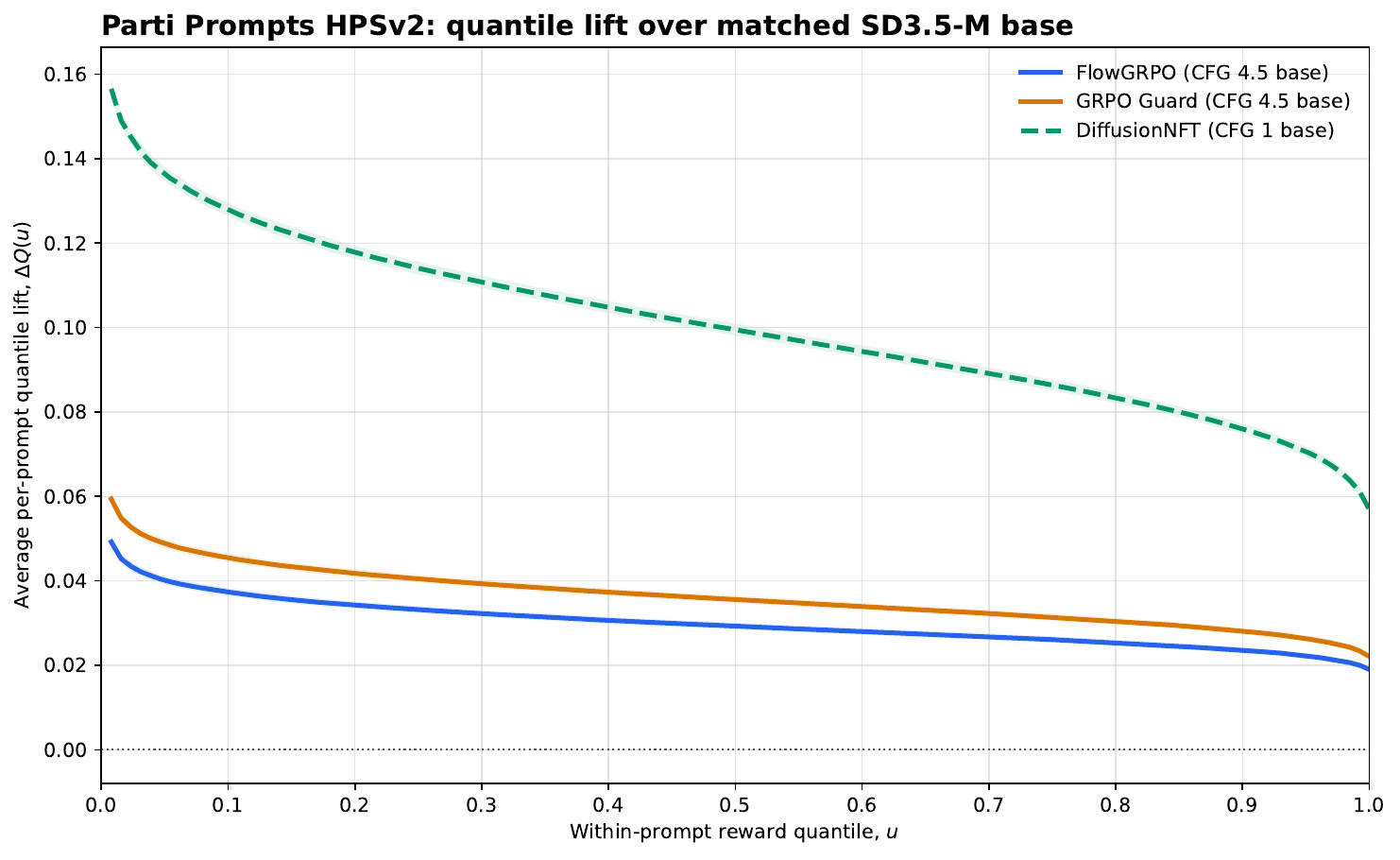}
        \caption{Parti-Prompts.}
        \label{fig:parti_quantile_lift}
    \end{subfigure}

    \caption{\textbf{Quantile transfer across held-out tasks.}
    GenEval2 exhibits negative upper-tail lift and frontier reversal; TextInVision exhibits positive but strongly decaying upper-tail lift; and Parti-Prompts with HPSv2 exhibits largely persistent positive lift.}
    \label{fig:quantile_lift_tasks}
\end{figure}

We estimate $Q_\pi(u\mid c)$ from the order statistics of the $G=128$ rewards per prompt, compute paired post-trained-minus-base quantile differences, and average across prompts; 95\% confidence intervals are obtained by paired prompt bootstrap. Figure~\ref{fig:quantile_lift_tasks} closely matches the regimes predicted by Theorem~\ref{thm:bon_quantile}. On GenEval2, lift becomes negative in the upper tail, explaining the eventual frontier reversal. On TextInVision, lift remains positive but contracts sharply near $u=1$, matching positive but decaying scaling transfer. On Parti-Prompts, upper-tail lift remains positive and largely stable, consistent with approximately persistent transfer. Thus, the shape of $\Delta Q(u)$ directly explains how Best-of-$1$ gains evolve with inference compute.

\subsection{Prompt Difficulty and Scaling Transfer}
\label{sec:atomicity_transfer}

For complex prompts, strong solutions may occupy only the extreme upper reward tail. Improving typical generations can therefore raise average reward without preserving the rare outcomes that additional sampling exploits.

\begin{figure}[t]
    \centering
    \includegraphics[width=0.8\linewidth]{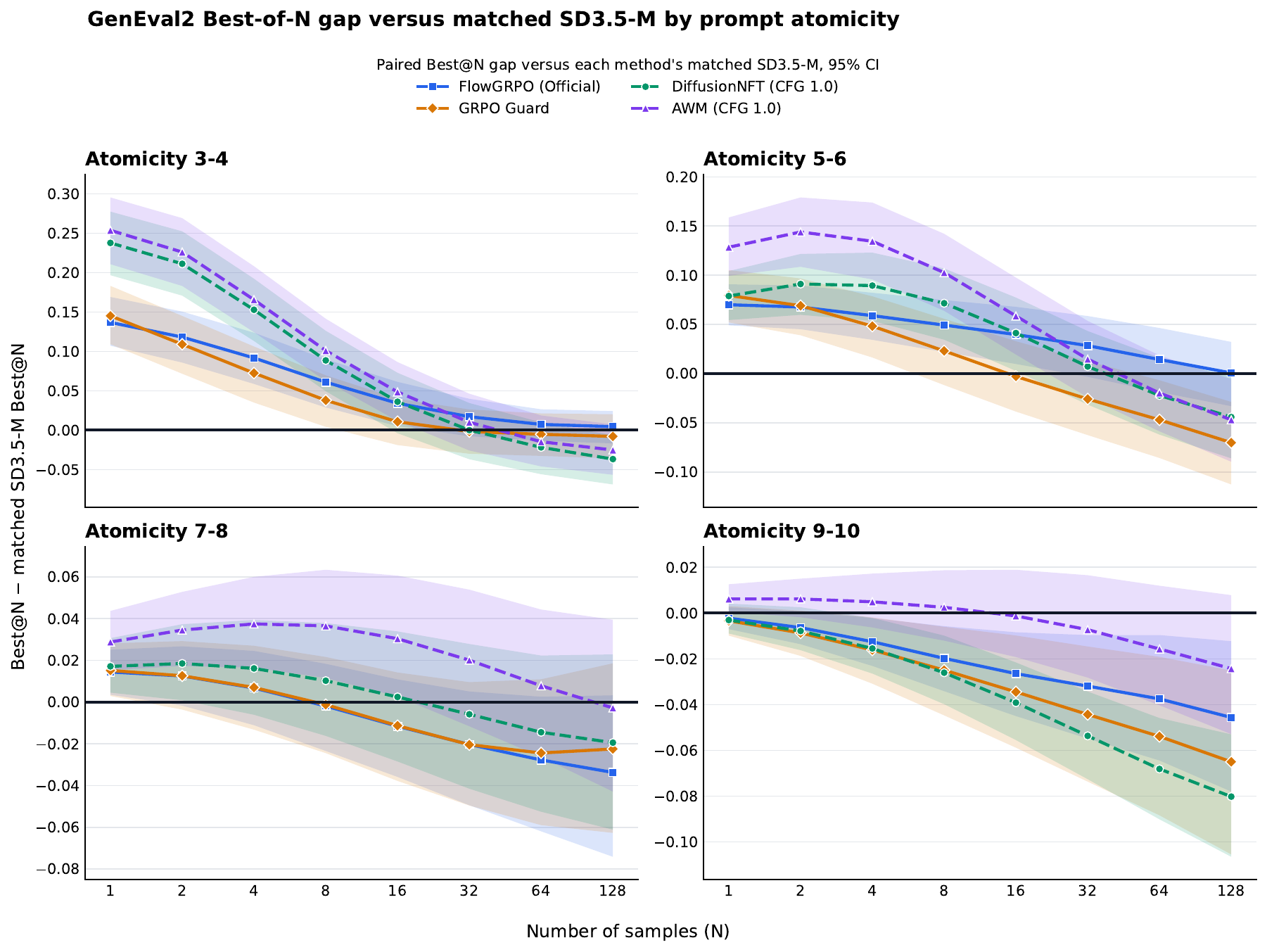}
    \caption{\textbf{Scaling transfer by prompt atomicity on GenEval2.}
    Best-of-$N$ advantages relative to matched SD3.5-M baselines: CFG 4.5 for FlowGRPO and GRPO-Guard, and CFG 1.0 for DiffusionNFT and AWM. Lower-atomicity prompts exhibit substantial low-budget gains, whereas the highest-atomicity group exhibits negative high-budget transfer across methods. Shading denotes 95\% confidence intervals; vertical scales differ across panels.}
    \label{fig:atomicity_transfer}
\end{figure}

Figure~\ref{fig:atomicity_transfer} stratifies GenEval2 by atomicity. On simpler prompts (atomicity 3--4), all four methods substantially improve Best-of-$1$, but these gains largely vanish as $N$ grows. On the most complex prompts (atomicity 9--10), Best-of-$1$ changes little, yet all four methods underperform their matched baselines at $N=128$. As Theorem~\ref{thm:bon_quantile} shows, larger inference budgets emphasize upper reward quantiles, exposing losses obscured at Best-of-$1$. Together with the quantile-lift results, this suggests that post-training can improve typical outcomes while weakening rare solutions needed for difficult compositions.

\section{Mitigating Scaling-Transfer Failure}
\label{sec:qupo}

\subsection{Controlling Quantile Utility}
\label{sec:qupo_objective}

Our analysis suggests that improving scaling transfer requires controlling which reward quantiles are emphasized during training. Consider the quantile-weighted objective
\begin{equation}
\mathcal{J}_{\phi}(\theta)
=
\E_c\!\left[
\int_0^1 Q_{\pi_\theta}(u\mid c)\phi(u)\,du
\right].
\label{eq:quantile_utility}
\end{equation}
Expected-reward optimization uses the uniform weighting $\phi(u)=1$, while Theorem~\ref{thm:bon_quantile} shows that Best-of-$k$ corresponds to $\phi_k(u)=ku^{k-1}$, increasingly emphasizing the upper tail as $k$ grows.

We therefore optimize a weighted combination of deployment budgets,
\begin{equation}
\phi_q(u)
=
\sum_{k=1}^{T} q_k k u^{k-1},
\qquad
\mathcal{J}_{\mathrm{QUPO}}(\theta)
=
\sum_{k=1}^{T} q_k B_k^{\pi_\theta}.
\label{eq:qupo_objective}
\end{equation}
In practice, we parameterize the budget weights as $q_k = {k^{\alpha}}$, giving two intuitive controls. The horizon $T$ determines the largest Best-of-$k$ utility explicitly represented during training, while $\alpha$ controls the relative emphasis placed on smaller versus larger sampling budgets. Notably, setting $T=1$ recovers the standard expected-reward optimization objective.

We call the resulting objective \textbf{Quantile Utility Policy Optimization (QUPO)}. QUPO does not directly regularize diversity or optimize only a single large-$k$ operating point; instead, it exposes the quantile weighting induced by test-time sampling as a controllable training objective.

\subsection{Practical Advantage Estimation}
\label{sec:qupo_advantage}

Given a rollout group of $G$ generations with rewards $r_1,\ldots,r_G$, we measure the contribution of sample $i$ to Best-of-$k$ utility as
\begin{equation}
a_i^{(k)}
=
\E_S\!\left[\max_{j\in S}r_j \mid i\in S,\ |S|=k\right]
-
\E_S\!\left[\max_{j\in S}r_j \mid i\notin S,\ |S|=k\right].
\label{eq:qupo_single_k_advantage}
\end{equation}
This measures how much including sample $i$ improves the expected Best-of-$k$ reward. Combining budgets gives the QUPO advantage
\begin{equation}
A_i^{\mathrm{QUPO}}
=
\sum_{k=1}^{T} kq_k a_i^{(k)}.
\label{eq:qupo_advantage}
\end{equation}
For $T\leq G-1$, this yields an unbiased policy-gradient estimator for Eq.~\ref{eq:qupo_objective}; the full derivation and discussion of its efficient computation are given in Appendix~\ref{app:qupo_policy_gradient}.

QUPO is a drop-in replacement for the scalar rollout advantage. In FlowGRPO~\cite{liu2026flow} and UniGRPO~\cite{liu2026unigrpo}, we replace the original group-relative advantage with $A_i^{\mathrm{QUPO}}$ while leaving the remaining optimization pipeline---including rollout generation, likelihood computation, clipping, and regularization---unchanged.

\begin{figure}[t]
    \centering
    \includegraphics[width=0.6\linewidth]{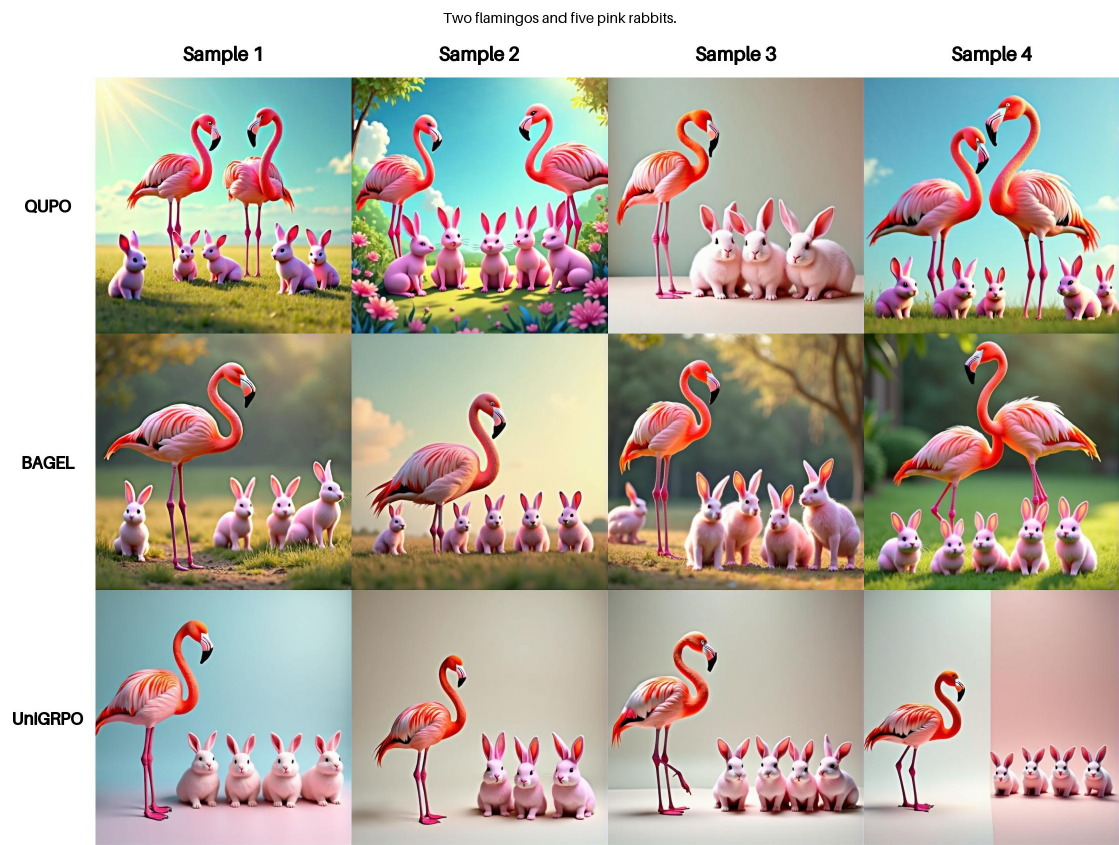}
    \caption{\textbf{Multiple generations for the same held-out GenEval2 prompt.}
Each row shows four independent samples from QUPO, the base BAGEL model, and UniGRPO for the prompt ``two flamingos and five pink rabbits'', under matched generation seeds. QUPO more consistently produces samples satisfying the requested composition, qualitatively reflecting its strong scaling transfer.}
    \label{fig:bagel_qualitative}
\end{figure}

\section{Experiments}
\label{sec:experiments}


\subsection{Experimental Setup}
\label{sec:exp_setup}

We focus on compositional generation, as it is the only diagnostic setting exhibiting frontier reversal. Models are post-trained on GenEval~\cite{ghosh2023geneval} and evaluated on the harder held-out GenEval2~\cite{kamath2025geneval} and ConceptMix~\cite{wu2024conceptmix} benchmarks.

We evaluate QUPO on SD3.5-M and BAGEL~\cite{deng2025emerging}. For SD3.5-M, we compare against FlowGRPO~\cite{liu2026flow}, DiffusionNFT~\cite{zheng2026diffusionnft}, GRPO-Guard~\cite{wang2026grpo}, and AWM~\cite{xue2025advantage}, generating 128 images per prompt; for BAGEL, we compare against UniGRPO~\cite{liu2026unigrpo}, generating 64 images per prompt to mitigate compute costs. QUPO changes only the rollout advantage in the corresponding training pipeline, with applicable training settings matched. We evaluate average Best-of-$N$ performance using the estimator from Sec.~\ref{sec:diagnostic_protocol}, ensuring standardized generation seeds for fair comparison across models. GenEval2 (800 prompts) and ConceptMix (2100 prompts) are scored with Soft-TIFA using Qwen3-VL-8B-Instruct~\cite{bai2025qwen3}, following established evaluation protocol \cite{kamath2025geneval}. Further implementation details, ablations, and diversity analysis are deferred to the appendix.

\subsection{Scaling Transfer on SD3.5-M}
\label{sec:sd35_results}

\begin{figure*}[t]
    \centering
    \begin{subfigure}[t]{0.61\textwidth}
        \centering
        \includegraphics[width=\linewidth]{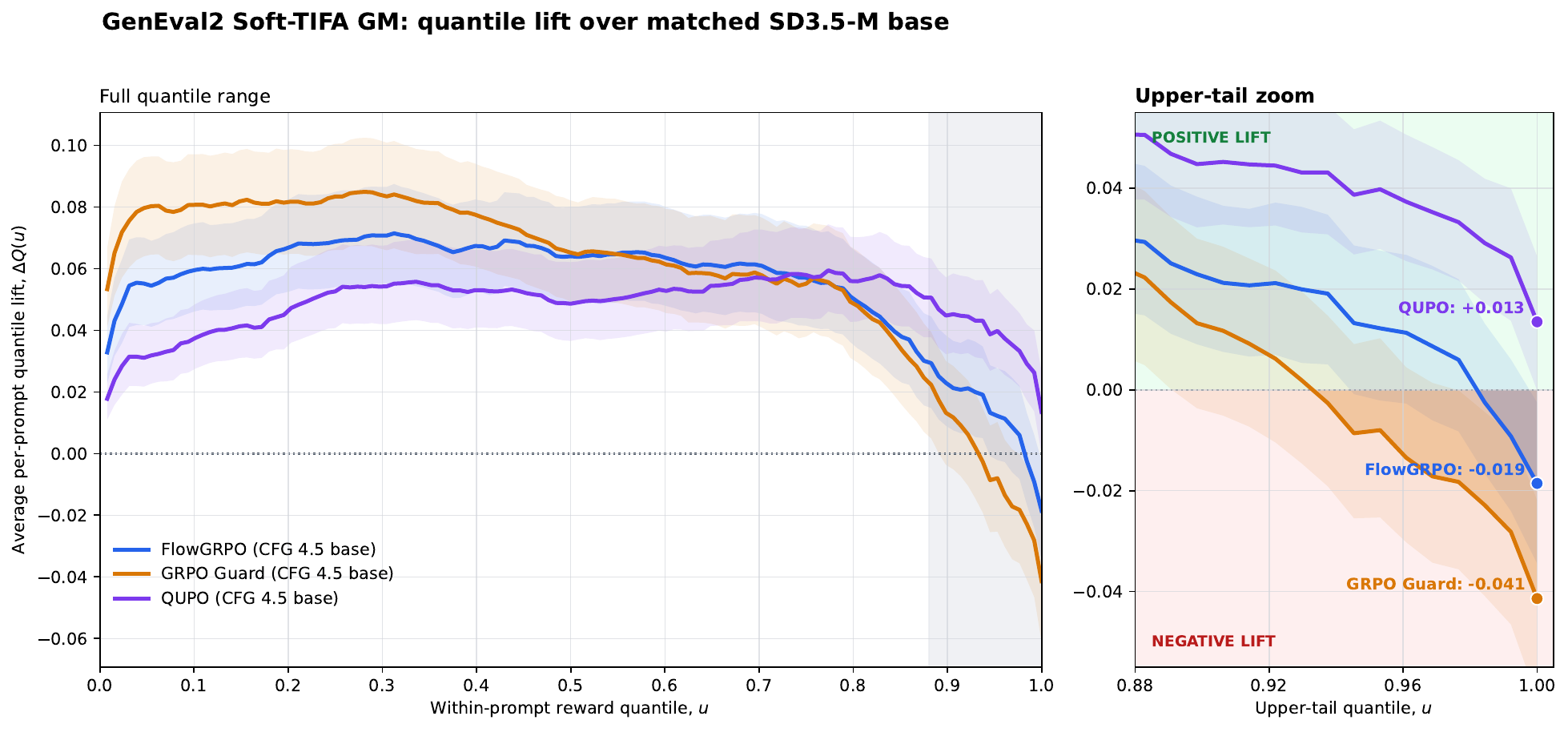}
        \caption{Reward quantile lift.}
    \end{subfigure}
    \hfill
    \begin{subfigure}[t]{0.37\textwidth}
        \centering
        \includegraphics[width=\linewidth]{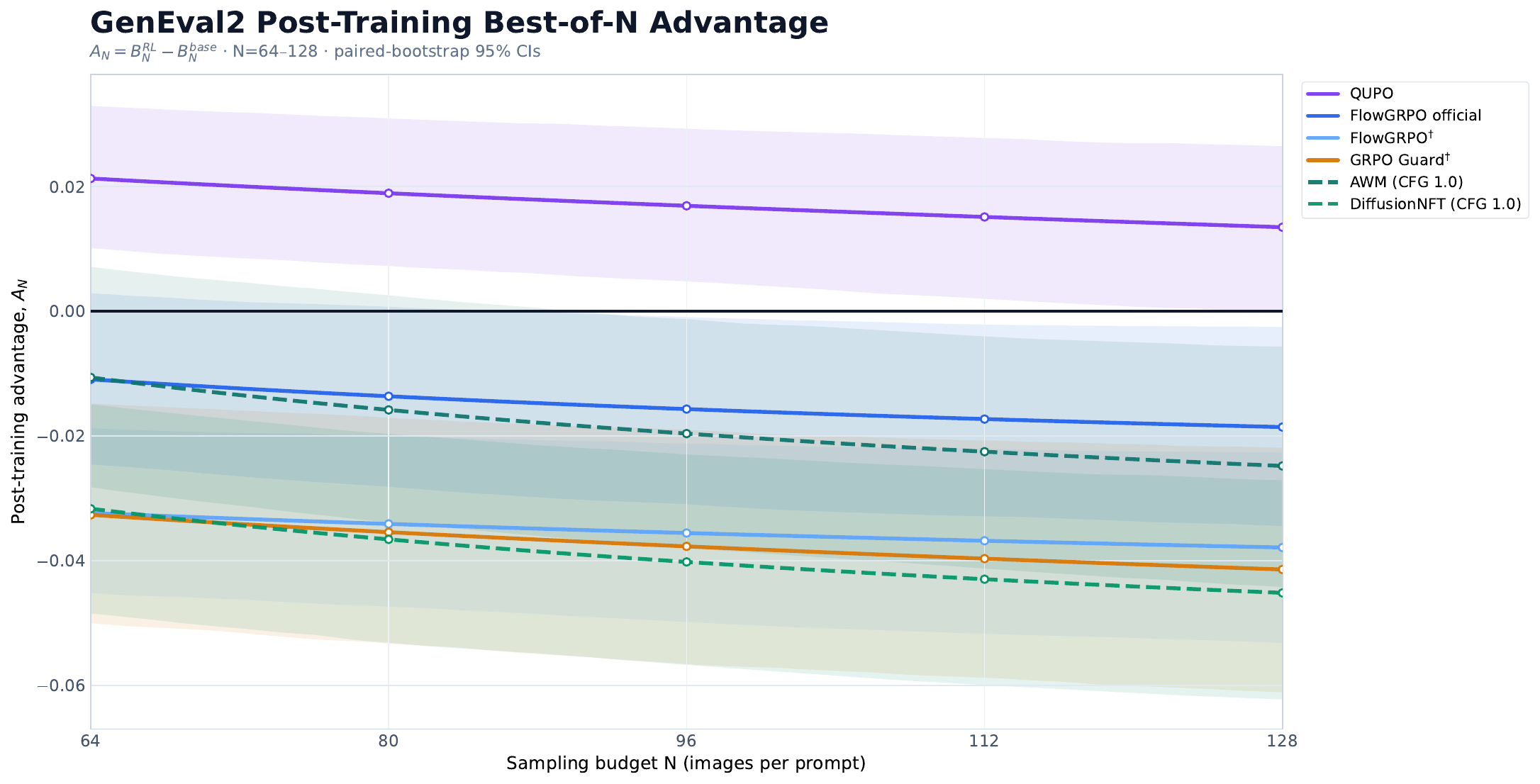}
        \caption{Large-$N$ scaling advantage.}
    \end{subfigure}

    \caption{\textbf{QUPO preserves upper-tail reward quality and mitigates scaling-transfer failure on GenEval2.}
    \textbf{Left:} FlowGRPO and GRPO-Guard exhibit negative upper-tail quantile lift relative to the matched base model, whereas QUPO remains positive.
    \textbf{Right:} Best-of-$N$ post-training advantage relative to the matched base model for $N=64$--$128$, with shaded bands indicating paired prompt-bootstrap 95\% confidence intervals. QUPO maintains positive estimated advantage, whereas standard RL methods exhibit frontier reversal.}
    \label{fig:qupo_quantile_lift}
\end{figure*}

We first evaluate QUPO on SD3.5-M using the GenEval post-training setup from our diagnostic study. On held-out GenEval2 prompts (Table~\ref{tab:sd35_main_results}), QUPO improves Best-of-$1$ from $0.1804$ to $0.2287$ and remains above the base model through Best-of-$128$ ($0.7000$ vs.\ $0.6850$). In contrast, FlowGRPO and GRPO-Guard achieve comparable or stronger Best-of-$1$ and higher in-distribution GenEval scores, yet are overtaken as $N$ increases. Figure~\ref{fig:qupo_quantile_lift} explains this difference: standard RL exhibits negative upper-tail quantile lift, whereas QUPO maintains positive lift in the region increasingly emphasized at large $N$.

\definecolor{qupoblue}{RGB}{230,242,255}

\begin{table*}[t]
\centering
\begin{subtable}[t]{0.72\textwidth}
\centering
\caption{GenEval2.}
\label{tab:sd35_qupo}
\resizebox{\linewidth}{!}{
\begin{tabular}{lcccccccccc}
\toprule
Model & CFG & GenEval & Bo1 & Bo4 & Bo8 & Bo16 & Bo32 & Bo64 & Bo96 & Bo128 \\
\midrule
SD3.5-M
    & 4.5 & 0.63 & 0.1804 & 0.3365 & 0.4193 & 0.4978 & 0.5679 & 0.6295 & 0.6625 & 0.6850 \\
\rowcolor{qupoblue}
QUPO
    & 4.5 & 0.88 & 0.2287 & \textbf{0.3841} & \textbf{0.4618} & \textbf{0.5333} & \textbf{0.5962} & \textbf{0.6508} & \textbf{0.6794} & \textbf{0.7000} \\
FlowGRPO$^\dagger$
    & 4.5 & 0.91 & 0.2190 & 0.3469 & 0.4135 & 0.4793 & 0.5412 & 0.5971 & 0.6269 & 0.6471 \\
FlowGRPO (official)
    & 4.5 & \textbf{0.95} & 0.2351 & 0.3727 & 0.4413 & 0.5067 & 0.5662 & 0.6186 & 0.6468 & 0.6664 \\
GRPO-Guard$^\dagger$
    & 4.5 & \textbf{0.95} & \textbf{0.2395} & 0.3644 & 0.4279 & 0.4883 & 0.5448 & 0.5968 & 0.6248 & 0.6436 \\
\midrule
SD3.5-M
    & 1.0 & 0.23 & 0.0998 & 0.2383 & 0.3223 & 0.4051 & 0.4813 & 0.5485 & 0.5833 & 0.6060 \\
DiffusionNFT
    & 1.0 & \textbf{0.95} & 0.1825 & 0.2991 & 0.3584 & 0.4152 & 0.4682 & 0.5168 & 0.5431 & 0.5608 \\
AWM
    & 1.0 & 0.90 & 0.2041 & 0.3240 & 0.3830 & 0.4391 & 0.4907 & 0.5379 & 0.5637 & 0.5812 \\
\bottomrule
\end{tabular}}
\vspace{1mm}

{\scriptsize
$^\dagger$ Trained with the official implementation using an optimization budget matched to QUPO.
}
\end{subtable}
\hfill
\begin{subtable}[t]{0.27\textwidth}
\centering
\caption{ConceptMix.}
\label{tab:conceptmix_qupo}
\resizebox{\linewidth}{!}{
\begin{tabular}{c|c>{\columncolor{qupoblue}}cc}
\toprule
$N$ & SD3.5-M & QUPO & FlowGRPO \\
\midrule
1   & 0.4107 & \textbf{0.4578} & 0.4546 \\
8   & 0.6742 & \textbf{0.6980} & 0.6704 \\
16  & 0.7338 & \textbf{0.7506} & 0.7199 \\
32  & 0.7823 & \textbf{0.7923} & 0.7607 \\
64  & 0.8220 & \textbf{0.8255} & 0.7947 \\
96  & 0.8411 & \textbf{0.8432} & 0.8119 \\
128 & 0.8543 & \textbf{0.8559} & 0.8231 \\
\bottomrule
\end{tabular}}
\end{subtable}

\caption{\textbf{QUPO improves scaling transfer on SD3.5-M.}
On GenEval2, standard RL methods eventually fall below the matched base model, whereas QUPO remains above it through Best-of-$128$. The improvement transfers to ConceptMix.}
\label{tab:sd35_main_results}
\end{table*}

While Theorem 1 provides an exact relationship between quantile lift and Best-of-N advantage within a fixed prompt distribution, an important question is whether upper-tail behavior measured on one benchmark can also characterize scaling transfer on another. We investigate this on ConceptMix using the quantile analysis obtained independently from GenEval2, without using ConceptMix reward distributions to construct the quantile lifts. As shown in Table \ref{tab:conceptmix_qupo}, the qualitative scaling regimes transfer across benchmarks: FlowGRPO exhibits negative upper-tail lift on GenEval2 and undergoes frontier reversal on ConceptMix, while QUPO maintains positive upper-tail lift on GenEval2 and approximately preserves the base model's large-N frontier on ConceptMix (0.8559 vs. 0.8543). This cross-benchmark consistency suggests that the upper-tail effects characterized by our quantile analysis can generalize across compositional prompt distributions, providing a useful diagnostic of scaling-transfer behavior beyond the benchmark on which the quantile shifts are measured.


\subsection{Scaling Transfer on BAGEL}
\label{sec:bagel_results}

We next test whether both the scaling-transfer failure and the benefits of QUPO extend beyond flow-based image generators. We use BAGEL~\cite{deng2025emerging}, a unified multimodal model with a qualitatively different generative architecture, and compare the base model against UniGRPO~\cite{liu2026unigrpo} and QUPO. All models are post-trained on GenEval and evaluated on held-out GenEval2 prompts at $512\times512$ resolution.

\begin{table}[t]
\centering
\caption{\textbf{Scaling transfer on BAGEL.}
UniGRPO improves Best-of-$1$ but degrades the large-$N$ frontier; QUPO largely preserves it.}
\label{tab:bagel_qupo}
\resizebox{\linewidth}{!}{
\begin{tabular}{lccccccc}
\toprule
Model & GenEval & Bo1 & Bo4 & Bo8 & Bo16 & Bo32 & Bo64 \\
\midrule
BAGEL
    & 0.74 & 0.2351 & 0.4188 & 0.5091 & 0.5890 & 0.6579 & {0.7166}\\
UniGRPO
    & \textbf{0.85} & 0.2420 & 0.4063 & 0.4892 & 0.5653 & 0.6321 & 0.6891 \\
\rowcolor{qupoblue}
QUPO
    & 0.83 & \textbf{0.2468} & \textbf{0.4240} & \textbf{0.5120} & \textbf{0.5905} & \textbf{0.6583} & \textbf{0.7177} \\
\bottomrule
\end{tabular}}
\end{table}

The same scaling-transfer failure appears on BAGEL: UniGRPO improves GenEval and GenEval2 Best-of-$1$ but falls increasingly below the base model as $N$ grows. QUPO performs best through Best-of-$64$, remaining slightly ahead of the base while outperforming UniGRPO by nearly $0.03$. Thus, both weakened scaling transfer and its mitigation through upper-tail-aware training generalize across qualitatively different visual generators.

\section{Conclusion}

We study \textit{scaling transfer} in visual generative models: whether gains from RL persist as test-time sampling compute increases on difficult held-out prompts. Across multiple RL methods, tasks, and model families, we show that stronger Best-of-$1$ performance does not necessarily imply a better quality--compute frontier, with regimes ranging from persistent gains to attenuation and complete frontier reversal. We explain these behaviors through a quantile-transfer analysis showing that larger Best-of-$N$ budgets increasingly depend on the upper tail of the reward distribution. Motivated by this insight, we introduce QUPO, a simple quantile-aware objective that better preserves upper-tail quality and improves scaling transfer on both SD3.5-M and BAGEL. More broadly, our findings expose a limitation of single-sample evaluation and motivate treating the full quality–compute frontier as an essential measure of visual RL post-training. Strong single-sample evaluation need not transfer to strong test-time scalability.



\bibliographystyle{plainnat}
\bibliography{main}

\clearpage

\beginappendix

\appendix

\startcontents[appendix]
\section*{Appendix Contents}
\printcontents[appendix]{}{1}{
    \setcounter{tocdepth}{2}
}

\section{Proofs and Derivations}
\label{app:proofs}

\subsection{Proof of Theorem~\ref{thm:bon_quantile}}
\label{app:proof_theorem1}

\begin{proof}
Fix a prompt $c$ and model $\pi$. Let
\[
F_\pi(t\mid c)
=
\mathbb{P}\!\left(r(\rvx,c)\leq t\right),
\qquad
\rvx\sim\pi(\cdot\mid c),
\]
with quantile function $Q_\pi(u\mid c)$. Let
$\ru_1,\ldots,\ru_N \overset{\mathrm{i.i.d.}}{\sim}\mathrm{Uniform}(0,1)$.
By the inverse-transform representation,
\[
Q_\pi(\ru_i\mid c)
\overset{d}{=}
r(\rvx_i,c),
\qquad
\rvx_i\sim\pi(\cdot\mid c).
\]
Since $Q_\pi(\cdot\mid c)$ is nondecreasing,
\begin{equation}
\max_{1\leq i\leq N} Q_\pi(\ru_i\mid c)
=
Q_\pi\!\left(\max_{1\leq i\leq N}\ru_i \,\middle|\, c\right).
\end{equation}

Define the maximum order statistic
\[
\ru_{(N)}=\max_{1\leq i\leq N}\ru_i.
\]
For $u\in[0,1]$,
\[
\mathbb{P}(\ru_{(N)}\leq u)
=
\prod_{i=1}^N \mathbb{P}(\ru_i\leq u)
=
u^N,
\]
so $\ru_{(N)}\sim\mathrm{Beta}(N,1)$ with density
\[
p_N(u)=Nu^{N-1}.
\]
Therefore, the conditional Best-of-$N$ reward is
\begin{align}
B_N^\pi(c)
&=
\E_{\rvx_1,\ldots,\rvx_N\sim\pi(\cdot\mid c)}
\left[
\max_i r(\rvx_i,c)
\right] \\
&=
\E_{\ru_{(N)}}
\left[
Q_\pi(\ru_{(N)}\mid c)
\right] \\
&=
\int_0^1 Q_\pi(u\mid c)\,Nu^{N-1}\,du.
\end{align}
Averaging over prompts and using integrability to exchange the prompt expectation and integral gives
\begin{equation}
B_N^\pi
=
\int_0^1
\E_c\!\left[Q_\pi(u\mid c)\right]
Nu^{N-1}\,du.
\end{equation}

Applying this identity to $\pi_{\mathrm{RL}}$ and $\pi_{\mathrm{ref}}$ and subtracting,
\begin{align}
A_N
&=
B_N^{\pi_{\mathrm{RL}}}
-
B_N^{\pi_{\mathrm{ref}}} \\
&=
\int_0^1
\E_c
\left[
Q_{\pi_{\mathrm{RL}}}(u\mid c)
-
Q_{\pi_{\mathrm{ref}}}(u\mid c)
\right]
Nu^{N-1}\,du \\
&=
\int_0^1
\Delta Q(u)\,Nu^{N-1}\,du.
\end{align}
Since $Nu^{N-1}$ is the density of
$\mathrm{Beta}(N,1)$, this is equivalently
\begin{equation}
A_N
=
\E_{\ru_N\sim\mathrm{Beta}(N,1)}
\left[\Delta Q(\ru_N)\right],
\end{equation}
which proves the result.
\end{proof}

\subsection{Derivation of the QUPO Policy Gradient}
\label{app:qupo_policy_gradient}

We derive the policy-gradient estimator used for the QUPO objective
\begin{equation}
\mathcal{J}_{\mathrm{QUPO}}(\theta)
=
\sum_{k=1}^{T} q_k B_k^{\pi_\theta}.
\end{equation}
We first consider a single Best-of-$k$ objective. For $k$ independent generations
$\rvx_1,\ldots,\rvx_k\sim\pi_\theta(\cdot\mid c)$, define
\begin{equation}
M_k = \max_{1\leq i\leq k} r(\rvx_i,c).
\end{equation}
Applying the score-function identity gives
\begin{align}
\nabla_\theta B_k^{\pi_\theta}
&=
\E_c
\E_{\rvx_1,\ldots,\rvx_k}
\left[
M_k
\nabla_\theta
\log
\prod_{i=1}^{k}
\pi_\theta(\rvx_i\mid c)
\right] \\
&=
\E_c
\E_{\rvx_1,\ldots,\rvx_k}
\left[
M_k
\sum_{i=1}^{k}
\nabla_\theta\log\pi_\theta(\rvx_i\mid c)
\right].
\label{eq:bk_policy_gradient}
\end{align}

In practice, visual RL methods sample a group of $G\geq k$ generations per prompt. Rather than selecting a single size-$k$ subset, we average Eq.~\ref{eq:bk_policy_gradient} over all $\binom{G}{k}$ subsets. For
$S\subseteq\{1,\ldots,G\}$ with $|S|=k$, let
\begin{equation}
M_S=\max_{j\in S} r_j,
\qquad
r_j=r(\rvx_j,c).
\end{equation}
An unbiased estimator of the gradient is then
\begin{equation}
\widehat{\nabla_\theta B_k}
=
\frac{1}{\binom{G}{k}}
\sum_{\substack{S\subseteq[G]\\|S|=k}}
M_S
\sum_{i\in S}
\nabla_\theta\log\pi_\theta(\rvx_i\mid c).
\label{eq:bk_subset_gradient}
\end{equation}
Rearranging the sum by generation yields
\begin{equation}
\widehat{\nabla_\theta B_k}
=
\frac{k}{G}
\sum_{i=1}^{G}
u_i^{(k)}
\nabla_\theta\log\pi_\theta(\rvx_i\mid c),
\label{eq:bk_u_gradient}
\end{equation}
where
\begin{equation}
u_i^{(k)}
=
\frac{1}{\binom{G-1}{k-1}}
\sum_{\substack{S\subseteq[G]\\|S|=k,\;i\in S}}
M_S
\end{equation}
is the average Best-of-$k$ utility over subsets containing sample $i$. The factor $k/G$ follows from
$\binom{G-1}{k-1}/\binom{G}{k}=k/G$.

To improve credit assignment, we subtract a leave-one-out baseline defined by the expected Best-of-$k$ utility of subsets that exclude sample $i$:
\begin{equation}
v_i^{(k)}
=
\frac{1}{\binom{G-1}{k}}
\sum_{\substack{S\subseteq[G]\\|S|=k,\;i\notin S}}
M_S,
\qquad k\leq G-1.
\end{equation}
Because $v_i^{(k)}$ depends only on the other $G-1$ samples, it is independent of $\rvx_i$ conditional on those samples. Hence,
\begin{equation}
\E\left[
v_i^{(k)}
\nabla_\theta\log\pi_\theta(\rvx_i\mid c)
\right]
=0,
\end{equation}
so subtracting this baseline preserves the expected policy gradient. Defining
\begin{equation}
a_i^{(k)}
=
u_i^{(k)}-v_i^{(k)},
\end{equation}
we obtain
\begin{equation}
\widehat{\nabla_\theta B_k}
=
\frac{k}{G}
\sum_{i=1}^{G}
a_i^{(k)}
\nabla_\theta\log\pi_\theta(\rvx_i\mid c).
\label{eq:bk_advantage_gradient}
\end{equation}

Finally, by linearity of the gradient,
\begin{align}
\widehat{\nabla_\theta
\mathcal{J}_{\mathrm{QUPO}}}
&=
\sum_{k=1}^{T}
q_k\widehat{\nabla_\theta B_k} \\
&=
\frac{1}{G}
\sum_{i=1}^{G}
\left(
\sum_{k=1}^{T}
kq_k a_i^{(k)}
\right)
\nabla_\theta\log\pi_\theta(\rvx_i\mid c).
\end{align}
Thus, defining the QUPO advantage
\begin{equation}
A_i^{\mathrm{QUPO}}
=
\sum_{k=1}^{T}
kq_k a_i^{(k)},
\label{eq:qupo_advantage_appendix}
\end{equation}
gives the sample-level policy-gradient estimator used in Sec.~\ref{sec:qupo_advantage}. In our implementation, $T\leq G-1$, ensuring that the leave-one-out baseline is well-defined for every included budget. For FlowGRPO and UniGRPO, this scalar advantage replaces the original group-relative advantage while the remaining optimization surrogate is unchanged.

For each fixed $k$, $a_i^{(k)}$ coincides with the canonical
centered finite-batch Max@$k$ advantage used by
MaxPO~\cite{takashiro2026advantage}; we include the derivation for completeness
and to make the coefficients $kq_k$ in the QUPO mixture explicit. While the per-budget advantage coincides with that of MaxPO, our primary contribution is the systematic empirical characterization of scaling transfer in visual RL, with QUPO serving as a simple intervention to mitigate the observed failures.

\subsection{Implementation of the QUPO Advantage}
\label{app:qupo_implementation}

Although the definition of $a_i^{(k)}$ averages over all size-$k$ subsets, it can be computed efficiently without enumerating subsets. For a rollout group with rewards $r_1,\ldots,r_G$, we first sort the samples such that
\[
r_1 \leq r_2 \leq \cdots \leq r_G.
\]
Consider the sample at sorted position $j$. Among size-$k$ subsets containing $j$, its reward $r_j$ is the maximum when the remaining $k-1$ elements are chosen from the preceding $j-1$ samples. If instead a later sample $\ell>j$ is the maximum, the remaining $k-2$ elements must be selected from the $\ell-2$ samples preceding $\ell$ other than $j$. Hence,
\begin{equation}
u_j^{(k)}
=
\frac{
r_j \binom{j-1}{k-1}
+
\sum_{\ell=j+1}^{G}
r_\ell \binom{\ell-2}{k-2}
}{
\binom{G-1}{k-1}
}.
\label{eq:qupo_u_closed}
\end{equation}
Similarly, for subsets excluding $j$, a sample $\ell<j$ is the maximum in
$\binom{\ell-1}{k-1}$ subsets, whereas a sample $\ell>j$ is the maximum in
$\binom{\ell-2}{k-1}$ subsets. Therefore,
\begin{equation}
v_j^{(k)}
=
\frac{
\sum_{\ell=1}^{j-1}
r_\ell \binom{\ell-1}{k-1}
+
\sum_{\ell=j+1}^{G}
r_\ell \binom{\ell-2}{k-1}
}{
\binom{G-1}{k}
}.
\label{eq:qupo_v_closed}
\end{equation}
We then compute
\begin{equation}
a_j^{(k)} = u_j^{(k)}-v_j^{(k)},
\qquad
A_j^{\mathrm{QUPO}}
=
\sum_{k=1}^{T} kq_k a_j^{(k)},
\end{equation}
and map the resulting advantages back to the original rollout ordering. We use the convention $\binom{n}{m}=0$ when $m<0$ or $m>n$. After sorting, the sums in Eqs.~\ref{eq:qupo_u_closed}--\ref{eq:qupo_v_closed} can be evaluated with prefix and suffix accumulations, giving $O(TG)$ computation after the initial $O(G\log G)$ sort. Since rollout groups are small in our experiments, this overhead is negligible relative to generation and likelihood evaluation.

QUPO is implemented as a direct replacement for the scalar rollout advantage in existing visual RL pipelines. For FlowGRPO~\cite{liu2026flow}, we replace the original group-relative advantage with $A_i^{\mathrm{QUPO}}$ and otherwise retain the official training procedure. Likewise, for UniGRPO~\cite{liu2026unigrpo}, we substitute the QUPO advantage while leaving the remaining optimization pipeline unchanged. QUPO does not modify the model parameterization, rollout procedure, policy surrogate, clipping or regularization terms, or the method used to compute rollout likelihoods. This last point is important for visual RL, where likelihood estimation for diffusion, flow, or unified multimodal generators is itself a central algorithmic component. QUPO operates only on the reward-to-advantage transformation and is therefore compatible with the likelihood computation already provided by each underlying method.

\section{Connections to Related Objectives}
\label{app:related}

In this section, we discuss the two objectives most closely related to QUPO. TailRL~\cite{ramasubramanian2026tail} is closely connected at the objective level through its harmonic mixture of Best-of-$k$ utilities, while Multi-Axis Max@K \cite{onoda2026multi} focuses on Max@K-aware post-training for diversity optimization. We emphasize that QUPO is a simple intervention motivated by the broader, primary contribution of this work: identifying and analyzing scaling-transfer failure in visual RL.

\subsection{Connection to TailRL}
\label{app:tailrl_connection}

TailRL, which was developed concurrently and independently, is closely related to QUPO in its motivation to preserve rare high-reward outcomes under increased inference-time sampling. For rewards in $[0,1]$, TailRL shows that its finite-order objective can be written as
\begin{equation}
\mathcal{J}_{\mathrm{TailRL}}^{(T)}(\theta)
=
\sum_{k=1}^{T}
\frac{B_k^{\pi_\theta}-1}{k}.
\label{eq:tailrl_truncated}
\end{equation}
Thus, TailRL implicitly assigns harmonic weight $1/k$ to each Best-of-$k$ objective.

QUPO uses the more general objective
\begin{equation}
\mathcal{J}_{\mathrm{QUPO}}(\theta)
=
\sum_{k=1}^{T} q_k B_k^{\pi_\theta},
\qquad
q_k = k^\alpha.
\end{equation}
Setting $\alpha=-1$ gives
\begin{equation}
\mathcal{J}_{\mathrm{QUPO}}^{\alpha=-1}(\theta)
=
\sum_{k=1}^{T}
\frac{B_k^{\pi_\theta}}{k}.
\end{equation}
Letting $H_T=\sum_{k=1}^{T}1/k$ denote the $T$-th harmonic number,
\begin{equation}
\mathcal{J}_{\mathrm{TailRL}}^{(T)}(\theta)
=
\mathcal{J}_{\mathrm{QUPO}}^{\alpha=-1}(\theta)
-
H_T.
\label{eq:tailrl_qupo_equivalence}
\end{equation}
Therefore, for a fixed horizon $T$, TailRL is exactly the $\alpha=-1$ instance of QUPO up to an additive constant, and the two objectives have identical gradients. This equivalence underscores that the novelty of our work does not rest on the introduction of a new class of Best-of-k objectives. Rather, these objectives provide a convenient means to test and address the scaling-transfer failure identified through our visual RL experiments.

The distinction is primarily in how the objective is motivated and parameterized. TailRL derives the harmonic weighting from a tail-likelihood objective, whereas QUPO begins from the equivalence between quantile weighting and the Best-of-$N$ quality--compute frontier. This perspective treats the weighting over sampling budgets as a controllable design choice: $\alpha$ determines how strongly larger Best-of-$k$ utilities are emphasized, while $T$ determines the largest budget represented explicitly during training. TailRL corresponds to the particular choice $\alpha=-1$.

Our ablations highlight two practical findings distinct from TailRL: the harmonic choice $\alpha=-1$ need not be optimal for visual generation, and the optimization horizon $T$ need not be coupled to the rollout group size $G$. In practice, moderate $T<G$ can already provide sufficient upper-tail emphasis to improve scaling at substantially larger deployment budgets. We study both effects in Sec.~\ref{app:ablations}. 

\subsection{Connection to Multi-Axis Max@$K$}

Multi-Axis Max@$K$~\cite{onoda2026multi} also uses set-level utilities
to post-train visual generators, but addresses a different objective.
It formalizes diversity as coverage of predefined semantic modes (i.e., color and gender) and
maximizes, for each mode, the score of its best representative among a
group of samples before aggregating across modes. This allows different
samples to receive credit for covering different categories.

QUPO instead considers a scalar reward and is derived from the
Best-of-$N$ quality--compute frontier. It optimizes a weighted family of
Best-of-$k$ utilities across deployment budgets, with the goal of preserving
the reward upper tail that determines test-time scaling. Thus, Multi-Axis
Max@$K$ uses set-level credit assignment to improve \textit{which modes} are
represented within a sample set, whereas QUPO uses it to improve \textit{how
scalar reward quality scales} with inference compute. Our primary
contribution further lies in identifying and analyzing scaling-transfer
failure across visual RL methods, tasks, and model families.



\section{Experimental and Implementation Details}
\label{app:implementation}

\subsection{Diagnostic Experiments}
\label{app:diagnostic-details}

\begin{table}[t]
\centering
\caption{\textbf{Diagnostic scaling-transfer settings.}
\texttt{Training} denotes reproduction using the official implementation; \texttt{Official LoRA} denotes an author-released adapter.}
\label{tab:diagnostic_settings}
\resizebox{\linewidth}{!}{
\begin{tabular}{lllll}
\toprule
Domain & Method & Post-training source & Acquisition & Held-out evaluation \\
\midrule
Compositionality
    & FlowGRPO & GenEval & \texttt{Official LoRA} & GenEval2 \\
    & DiffusionNFT & GenEval & \texttt{Official training} & GenEval2 \\
    & GRPO-Guard & GenEval & \texttt{Official LoRA} & GenEval2 \\
    & AWM & GenEval & \texttt{Official training} & GenEval2 \\
\midrule
Text rendering
    & FlowGRPO & OCR training set & \texttt{Official LoRA} & TextInVision \\
    & DiffusionNFT & OCR training set & \texttt{Official training} & TextInVision \\
    & GRPO-Guard & OCR training set & \texttt{Official LoRA} & TextInVision \\
\midrule
Preference alignment
    & FlowGRPO & Pick-a-Pic + PickScore & \texttt{Official LoRA} & Parti-Prompts + HPSv2 \\
    & DiffusionNFT & Pick-a-Pic + PickScore & \texttt{Official training} & Parti-Prompts + HPSv2 \\
    & GRPO-Guard & Pick-a-Pic + PickScore & \texttt{Official LoRA} & Parti-Prompts + HPSv2 \\
\bottomrule
\end{tabular}}
\end{table}

\textbf{Compositional Generation.} We evaluated SD3.5-M models post-trained on GenEval~\cite{ghosh2023geneval} prompts, using a GenEval detector-derived reward, following established protocol from FlowGRPO~\cite{liu2026flow}. The GenEval training prompts are open-sourced in the FlowGRPO codebase.

We evaluated these models on GenEval2~\cite{kamath2025geneval}, which consists of compositional text-to-image prompts with broader coverage of visual primitives and higher degrees of compositionality than GenEval, making it substantially more challenging for modern generators. 

We use the official Soft-TIFA evaluator~\cite{kamath2025geneval} with Qwen3-VL-8B-Instruct~\cite{bai2025qwen3} as the VQA scorer. Each GenEval2 prompt is associated with a set of atomic visual questions targeting its constituent objects, attributes, relations, and counts. For each generated image, Qwen3-VL is queried with these questions, and Soft-TIFA uses the model probability assigned to the correct answer as a soft per-atom score. We use the standard Soft-TIFA GM aggregation, taking the geometric mean of the atom-level scores to obtain the reward for each image.

\textbf{Text Rendering.} We evaluated SD3.5-M models post-trained using the open-sourced OCR training prompts and reward from the official FlowGRPO codebase: PaddleOCR is applied to each generated image, and all recognized text is concatenated, lowercased, and stripped of whitespace. Let $y$ denote the target text extracted from the prompt and $\hat y$ the recognized text. The reward is based on normalized Levenshtein similarity,
\begin{equation}
r_{\mathrm{OCR}}
=
1-\frac{\min\{d_{\mathrm{Lev}}(\hat y,y),\,|y|\}}{|y|},
\end{equation}
with reward $1$ assigned whenever $y$ appears as a substring of $\hat y$. This yields a score in $[0,1]$, where $1$ corresponds to perfect text rendering. 

For evaluation on TextInVision~\cite{fallah2025textinvision}, we use the same normalized-Levenshtein OCR scoring procedure on the benchmark's target text. We specifically evaluated on a subset of 800 prompts from TextInVision, which provides a substantially harder held-out OCR benchmark than the training set. Its target text averages 6.1 words and 36.4 characters—roughly twice the training averages—and includes long sentences, rare or synthetic words, numbers, and punctuation. Moreover, 90$\%$ of the prompts place text within complex scenes or constrained surfaces, and all targets and scene contexts were filtered against the training data to prevent exact or near-duplicate leakage.

\textbf{Preference Alignment.} We evaluated SD3.5-M models post-trained on the Pick-a-Pic v2~\cite{kirstain2023pick} prompt set, using PickScore as the training reward. Pick-a-Pic contains user-written text-to-image prompts collected through a preference-data collection platform, together with human comparisons between generated images. 

To evaluate transfer beyond both the training prompts and reward model, we test on the held-out Parti Prompts benchmark and score each generated image with HPSv2~\cite{wu2023human}, a learned human-preference metric conditioned on the corresponding text prompt. We then compute Best-of-$N$ performance using the HPSv2 scores across the 128 generations for each prompt.

\subsection{Post-Training Hyperparameters}

We present the detailed hyperparameter configuration in Tables~\ref{tab:sd35_training_details} and~\ref{tab:bagel_training_details}. As shown, all applicable hyperparameters are matched between QUPO and the respective baseline, differing only in the per-sample advantage computation. FlowGRPO and UniGRPO use Equation~\ref{eq:grpo_adv} to compute the per-sample advantage, while QUPO uses Equation~\ref{eq:qupo_advantage}. 

\begin{table*}[t]
\centering
\setlength{\tabcolsep}{2.8pt}
\renewcommand{\arraystretch}{0.92}
\scriptsize

\begin{subtable}[t]{0.49\textwidth}
\centering
\caption{\textbf{SD3.5-M training.}}
\label{tab:sd35_training_details}
\resizebox{\linewidth}{!}{
\begin{tabular}{lcc}
\toprule
Setting & FlowGRPO$^\dagger$ & QUPO \\
\midrule
Base / reward & SD3.5-M / GenEval & Same \\
Resolution / GPUs & $512^2$ / 32 & Same \\
Rollout steps / CFG & 10 / 4.5 & Same \\
Group size $G$ & 24 & Same \\
\midrule
LoRA rank / $\alpha$ & 32 / 64 & Same \\
Learning rate & $3{\times}10^{-4}$ & Same \\
AdamW $(\beta_1,\beta_2)$ / WD
    & $(0.9,0.999)$ / $10^{-4}$ & Same \\
Per-GPU batch / grad. accum.
    & 9 / 2 & Same \\
Updates / epoch & 2 & Same \\
$\#$ Epochs & 3350 & Same \\
\midrule
PPO clip / KL & $10^{-4}$ / 0.04 & Same \\
Max grad. norm & 1.0 & Same \\
Trained timesteps & 9 / 10 & Same \\
Timestep fraction / noise & 0.99 / 0.7 & Same \\
Advantage clamp & [-5,5] & Same \\
\midrule
Advantage
    & Group-relative & QUPO \\
$T$ / $\alpha$ & -- & 16 / $-0.5$ \\
\bottomrule
\end{tabular}}
\end{subtable}
\hfill
\begin{subtable}[t]{0.49\textwidth}
\centering
\caption{\textbf{BAGEL training.}}
\label{tab:bagel_training_details}
\resizebox{\linewidth}{!}{
\begin{tabular}{lcc}
\toprule
Setting & UniGRPO & QUPO \\
\midrule
Base / reward & BAGEL / GenEval & Same \\
Resolution / GPUs & $512^2$ / 64 & Same \\
Rollout steps / CFG & 25 / 1.0 & Same \\
Group size $G$ & 24 & Same \\
Images / reasoning chain & 1 & Same \\
\midrule
AR sampling
    & $T{=}1.0,\ p{=}0.9,\ k{=}1024$ & Same \\
Max reasoning tokens & 1024 & Same \\
Fine-tuning & Full (no LoRA) & Same \\
LR (reason / image)
    & $10^{-6}$ / $3{\times}10^{-5}$ & Same \\
AdamW $(\beta_1,\beta_2)$ / WD
    & $(0.9,0.999)$ / 0 & Same \\
Batch / grad. accum.
    & 1 / 12 & Same \\
Effective batch / updates
    & 672 / 2 & Same \\
Inner epochs & 1 & Same \\
$\#$ Epochs & 150 & Same \\
\midrule
PPO clip (text / image)
    & $10^{-2}$ / $10^{-6}$ & Same \\
KL (text) & 0 & Same \\
Velocity-MSE / ratio norm.
    & $1.5{\times}10^{-5}$ / on & Same \\
Max grad. norm & 1.0 & Same \\
Trained timesteps & 3 / 25 & Same \\
Timestep fraction / noise
    & $[0,0.2]$ / 0.8 & Same \\
Advantage clamp & [-5,5] & Same \\
\midrule
Advantage
    & Group-relative & QUPO \\
$T$ / $\alpha$ & -- & 16 / $-0.875$ \\
\bottomrule
\end{tabular}}
\end{subtable}

\caption{\textbf{Post-training hyperparameters for SD3.5-M and BAGEL.}
Within each backbone, QUPO uses the same optimization pipeline as the corresponding baseline and changes only the rollout advantage. ($\dagger$: FlowGRPO trained at matched budget, not the official checkpoint.)}
\label{tab:training_details}
\end{table*}

\subsection{Evaluation Hyperparameters}

We generated 128 images per prompt for SD3.5-M evaluation and 64 images per prompt for BAGEL, using matched seeds across methods to ensure fair comparison. The detailed inference hyperparameters are contained in Table \ref{tab:inference_details}. The hyperparameters used for SD3.5-M evaluation are shared across both the diagnostic experiments and the QUPO experiments.

\begin{table}[t]
\centering
\caption{\textbf{Inference hyperparameters.}
All Best-of-$N$ evaluations use the same inference configuration within each backbone.}
\label{tab:inference_details}
\scriptsize
\setlength{\tabcolsep}{4pt}
\begin{tabular}{lcc}
\toprule
Setting & SD3.5-M & BAGEL \\
\midrule
Resolution & $512\times512$ & $512\times512$ \\
Generation steps & 40 & 50 \\
CFG & 4.5$^\dagger$ & 4.0 / 1.5 (text / image) \\
Flow shift & -- & 3.0 \\
AR temperature & -- & 0.3 \\
Max reasoning tokens & -- & 1024 \\
$\#$ Samples per prompt & 128 & 64 \\
\bottomrule
\end{tabular}

\vspace{1mm}
\footnotesize{
$^\dagger$ DiffusionNFT and AWM are evaluated with CFG $=1.0$ following their corresponding inference settings; FlowGRPO, GRPO-Guard, and QUPO use CFG $=4.5$.
}
\end{table}

\section{Additional Ablations}
\label{app:ablations}

\subsection{Effect of the Optimization Horizon $T$}
\label{app:t_ablation}

We study the effect of the QUPO horizon $T$, which controls the largest Best-of-$k$ utility explicitly included in the training objective. We fix $\alpha=-0.5$ and train otherwise identical SD3.5-M models to the same optimization step. Results are shown in Table~\ref{tab:t_ablation}, alongside the pretrained SD3.5-M base model for reference.

\begin{table}[t]
\centering
\caption{\textbf{Effect of the QUPO horizon $T$.}
All post-trained models use $\alpha=-0.5$ and are evaluated at training step 3350. While $T=1$ improves single-sample quality, it degrades large-$N$ performance relative to the base model. Moderate horizons improve the full held-out Best-of-$N$ frontier, with performance peaking around $T=16$.}
\label{tab:t_ablation}
\scriptsize
\setlength{\tabcolsep}{3.5pt}
\begin{tabular}{c|c|c|cccccccc}
\toprule
$T$ & Step & GenEval & Bo1 & Bo4 & Bo8 & Bo16 & Bo32 & Bo64 & Bo96 & Bo128 \\
\midrule
Base
& -- & 0.63 & 0.1804 & 0.3365 & 0.4193 & 0.4978 & 0.5679 & 0.6295 & 0.6625 & 0.6850 \\
\midrule
1
& 3350 & 0.79 & 0.2068 & 0.3551 & 0.4300 & 0.5010 & 0.5661 & 0.6237 & 0.6536 & 0.6729 \\
8
& 3350 & 0.87 & 0.2222 & 0.3751 & 0.4519 & 0.5239 & 0.5889 & 0.6465 & 0.6768 & 0.6965 \\
16
& 3350 & \textbf{0.88} & \textbf{0.2287} & \textbf{0.3841} & \textbf{0.4618} &
\textbf{0.5333} & \textbf{0.5962} & \textbf{0.6508} & \textbf{0.6794} & \textbf{0.7000} \\
23
& 3350 & 0.87 & 0.2256 & 0.3765 & 0.4523 & 0.5243 & 0.5899 & 0.6476 & 0.6775 & 0.6973 \\
\bottomrule
\end{tabular}
\end{table}

Increasing $T$ from $1$ to $8$ and $16$ consistently improves the held-out quality--compute frontier, while extending the horizon further to $T=23$ provides no additional benefit. Notably, optimizing only through $T=16$ improves performance even at much larger deployment budgets such as Best-of-$64$ and Best-of-$128$, suggesting that directly including very large $k$ is not necessary to obtain stronger upper-tail behavior.

Crucially, the $T=1$ setting recovers expected-reward optimization and still exhibits scaling-transfer failure. For $k=1$, the QUPO advantage reduces to
\begin{equation}
a_i^{(1)}
=
r_i-\frac{1}{G-1}\sum_{j\neq i}r_j
=
\frac{G}{G-1}(r_i-\bar r),
\end{equation}
which is proportional to the mean-centered reward used by FlowGRPO, but without its division by the within-group reward standard deviation. Thus, both objectives are aligned with Best-of-$1$, while FlowGRPO generally applies a more aggressively scaled update through Z-score normalization. Consistent with this difference, $T=1$ improves GenEval and Best-of-$1$ less than FlowGRPO, but nevertheless exhibits the same qualitative scaling failure: its Best-of-$1$ improves over the pretrained base ($0.2068$ vs.\ $0.1804$), while its improvement decays with $N$ and reverses by Best-of-$128$ ($0.6729$ vs.\ $0.6850$). In contrast, introducing multi-sample utility with $T=8$ or $T=16$ preserves a positive advantage throughout the measured frontier, reaching Best-of-$128$ scores of $0.6965$ and $0.7000$, respectively. Together, these results suggest that frontier reversal is not specific to FlowGRPO's Z-score advantage or aggressive reward scaling, but arises more fundamentally when post-training is aligned only with expected single-sample reward; explicitly incorporating Best-of-$k$ utilities mitigates this failure.


\subsection{Effect of the Weighting Parameter $\alpha$}
\label{app:alpha_ablation}

We next vary the QUPO weighting parameter $\alpha$ while fixing the optimization horizon to $T=16$. Recall that $q_k=k^\alpha$, so increasing $\alpha$ places relatively greater weight on larger Best-of-$k$ utilities. Table~\ref{tab:alpha_ablation} reports the resulting GenEval and held-out GenEval2 performance.

\begin{table}[t]
\centering
\caption{\textbf{Effect of the QUPO weighting parameter $\alpha$.}
All runs use $T=16$. The harmonic weighting $\alpha=-1$, corresponding to TailRL, is competitive but is not uniformly optimal; $\alpha=-0.5$ achieves the strongest performance across most sampling budgets.}
\label{tab:alpha_ablation}
\scriptsize
\setlength{\tabcolsep}{3.5pt}
\begin{tabular}{c|c|cccccccc}
\toprule
$\alpha$ & GenEval & Bo1 & Bo4 & Bo8 & Bo16 & Bo32 & Bo64 & Bo96 & Bo128 \\
\midrule
$-2.0$
& 0.79 & 0.2119 & 0.3661 & 0.4434 & 0.5160 & 0.5817 & 0.6401 & 0.6710 & 0.6914 \\
$-1.0$
& 0.85 & 0.2194 & 0.3676 & 0.4427 & 0.5153 & 0.5831 & 0.6452 & 0.6787 & {0.6920} \\
$-0.75$
& 0.85 & 0.2218 & 0.3734 & 0.4497 & 0.5216 & 0.5862 & 0.6427 & 0.6721 & 0.6911 \\
$-0.5$
& \textbf{0.88} & \textbf{0.2287} & \textbf{0.3841} & \textbf{0.4618} &
\textbf{0.5333} & \textbf{0.5962} & \textbf{0.6508} & \textbf{0.6794} & \textbf{0.7000} \\
\bottomrule
\end{tabular}
\end{table}

The results show that the harmonic choice $\alpha=-1$ is not uniquely optimal for visual generation. Increasing $\alpha$ from $-2$ toward $-0.5$ generally improves both in-distribution GenEval performance and most of the held-out quality--compute frontier, with $\alpha=-0.5$ performing best from Best-of-$1$ through Best-of-$128$.

This ablation supports treating the Best-of-$k$ weighting as a controllable design choice rather than fixing the harmonic weighting a priori. In particular, moderately stronger emphasis on larger $k$ can improve broad scaling performance, while the optimal weighting may depend on the range of inference budgets of interest.

\subsection{Effect of Normalized Advantage Coefficients}
\label{app:coef_ablation}

In the preceding ablation, varying $\alpha$ changes not only the relative weighting of different Best-of-$k$ utilities, but also the overall magnitude of the effective advantage coefficients $kq_k$. To disentangle these effects, we additionally study a normalized weighting scheme
\begin{equation}
    q_k = c_\alpha k^\alpha,
    \qquad
    c_\alpha = \frac{T}{\sum_{j=1}^{T} j^{\alpha+1}},
\end{equation}
which ensures that the effective coefficients $w_k=kq_k$ satisfy
$\frac{1}{T}\sum_{k=1}^{T}w_k=1$.
This keeps the total effective coefficient weight fixed across choices of $\alpha$, while preserving their relative emphasis on different sampling budgets.

\begin{table}[t]
\centering
\caption{\textbf{Effect of normalized advantage coefficients.}
All runs use $T=16$ with normalized weights satisfying
$\frac{1}{T}\sum_{k=1}^{T}kq_k=1$.
Increasing $\alpha$ shifts performance from smaller toward larger inference budgets, with $\alpha=0$ achieving the strongest Best-of-$8$ through Best-of-$128$ results.}
\label{tab:coef_ablation}
\scriptsize
\setlength{\tabcolsep}{3.5pt}
\begin{tabular}{c|cccccccc}
\toprule
$\alpha$ & Bo1 & Bo4 & Bo8 & Bo16 & Bo32 & Bo64 & Bo96 & Bo128 \\
\midrule
$-1.0$
& \textbf{0.2194} & \textbf{0.3676} & 0.4427 & 0.5153 &
0.5831 & 0.6452 & 0.6787 & 0.6920 \\
$-0.5$
& 0.2087 & 0.3653 & 0.4446 & 0.5196 &
0.5873 & 0.6457 & 0.6787 & 0.7003 \\
$0.0$
& 0.2048 & 0.3669 & \textbf{0.4489} & \textbf{0.5246} &
\textbf{0.5916} & \textbf{0.6496} & \textbf{0.6800} & \textbf{0.7012} \\
\bottomrule
\end{tabular}
\end{table}

Table~\ref{tab:coef_ablation} reports the results on held-out GenEval2 with $T=16$ fixed. Under coefficient normalization, increasing $\alpha$ from $-1$ to $0$ decreases Best-of-$1$ performance from $0.2194$ to $0.2048$, while improving Best-of-$128$ from $0.6920$ to $0.7012$. More generally, larger $\alpha$ consistently improves performance from Best-of-$8$ through Best-of-$128$, demonstrating a tradeoff between single-sample quality and performance at larger inference budgets.

These results further support the interpretation of QUPO as a controllable weighting of the quality--compute frontier. Even when the total effective coefficient weight is held fixed, placing greater relative emphasis on larger Best-of-$k$ utilities shifts the learned frontier toward higher sampling budgets. While this normalization does not directly equalize the variance of the resulting per-sample advantages, it removes the change in total coefficient weight as a confounding factor in the $\alpha$ ablation.

\subsection{Scaling-Transfer Robustness Across Evaluators and Selectors}

Our primary GenEval2 compositional evaluation uses \textit{oracle Best-of-$N$} selection, where candidates are selected using the same Qwen3 Soft-TIFA score used for evaluation. This allows us to characterize how RL post-training reshapes the per-prompt reward distribution without confounding effects from a mismatch between selection and evaluation rewards. To assess the robustness of our findings to the choice of reward model and selection procedure, we consider two additional GenEval2 settings: (i) oracle Best-of-$N$ using InternVL3.5-8B~\cite{wang2025internvl35} as the Soft-TIFA scorer, and (ii) non-oracle Best-of-$N$, where Qwen3.5-9B~\cite{qwen2026qwen35} serves as the Soft-TIFA selector and InternVL3.5-8B as the evaluator.

\begin{figure*}[t]
    \centering
    \begin{subfigure}[t]{0.48\textwidth}
        \centering
        \includegraphics[width=\linewidth]{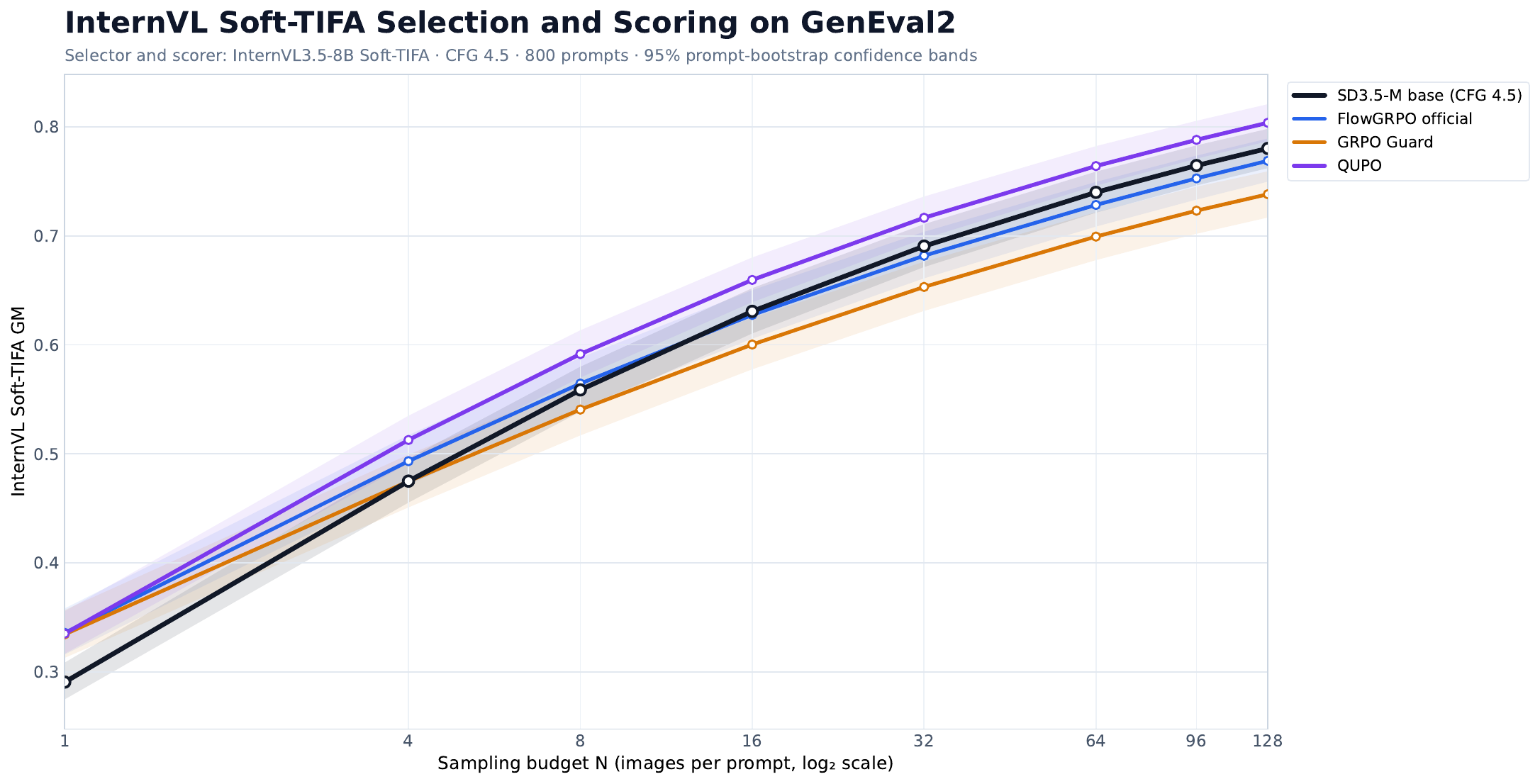}
        \caption{Oracle Best-of-$N$ evaluation on GenEval2 using InternVL3.5 Soft-TIFA scores.}
    \end{subfigure}
    \hfill
    \begin{subfigure}[t]{0.48\textwidth}
        \centering
        \includegraphics[width=\linewidth]{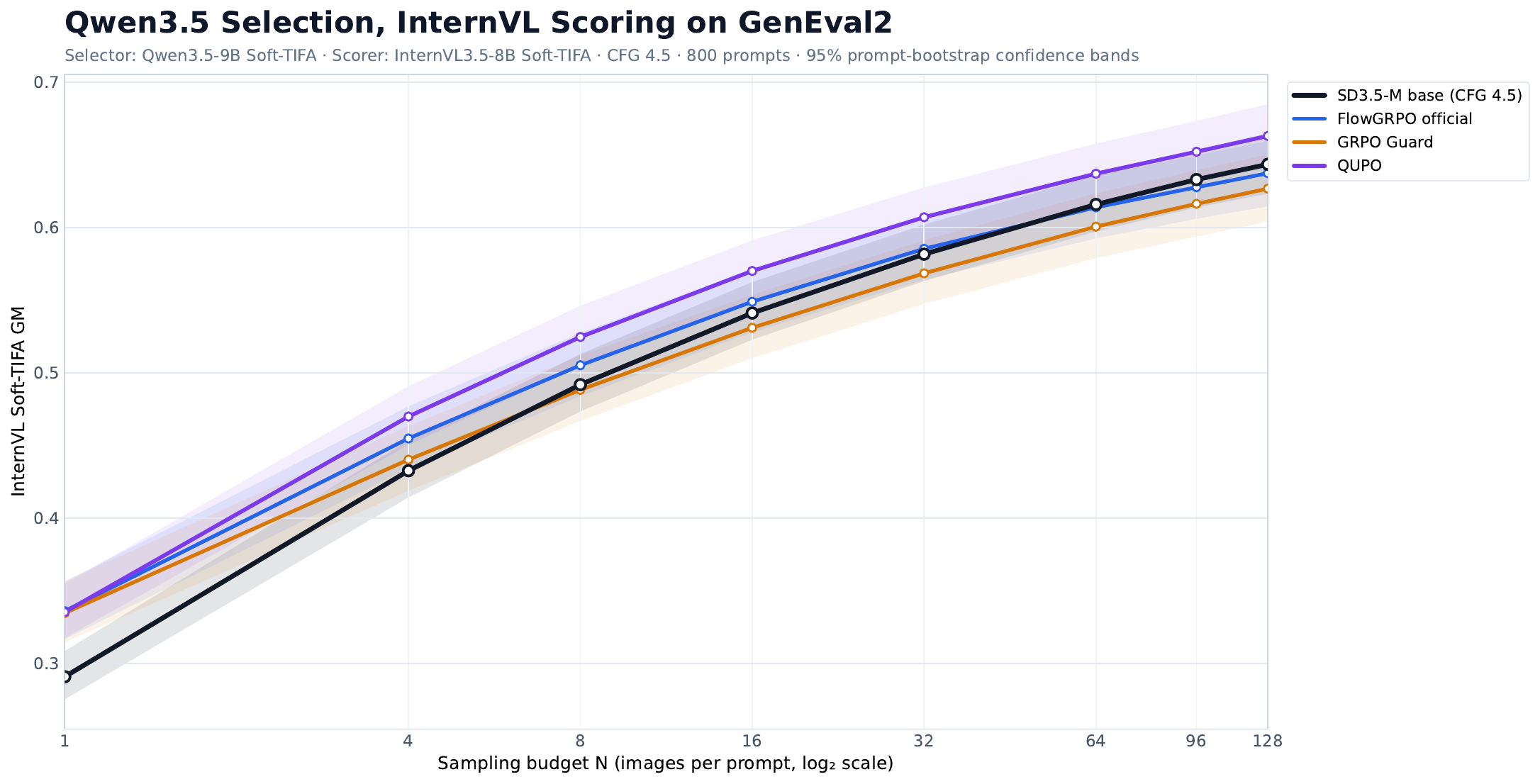}
        \caption{Non-oracle Best-of-$N$ evaluation using Qwen3.5 as the scorer and InternVL3.5 as the evaluator}
    \end{subfigure}

    \caption{Additional Best-of-$N$ evaluation on GenEval2 with different VLMs. As shown, the scaling transfer reversal occurs for FlowGRPO and GRPO Guard, while QUPO mitigates this failure.}
    \label{fig:extra_bon}
\end{figure*}

As shown in Figure \ref{fig:extra_bon}, scaling-transfer reversal persists under both settings. Although FlowGRPO and GRPO Guard improve Best-of-$1$ performance relative to the base SD3.5-M model, the advantage diminishes as $N$ increases and eventually becomes negative. This behavior occurs both when InternVL3.5 is used for selection and evaluation and when Qwen3.5 selects candidates to be evaluated by InternVL3.5. On the other hand, QUPO largely mitigates the scaling transfer failure in both cases, remaining above the RL models and base SD3.5-M as $N$ increases towards $128$. Together with our primary Qwen3-VL-based evaluation, these results demonstrate that the observed scaling-transfer failure is not specific to the choice of VLM evaluator or to using identical rewards for Best-of-$N$ selection and evaluation.

\subsection{Comparison with a Single Best-of-$k$ Objective}
\label{sec:single_budget}

To assess the benefit of mixing inference budgets, we compare QUPO with a
single Best-of-16 objective under matched training conditions. For this baseline, we retain only the Best-of-16 component and set its coefficient to the sum of all QUPO advantage coefficients, yielding $A_i = \left(\sum_{k=1}^{16} kq_k\right)a_i^{(16)} \approx 44.47a_i^{(16)}$.
As shown in Table~\ref{tab:single_budget}, the single-budget baseline
remains above the base model through Best-of-128, mitigating frontier
reversal. QUPO achieves higher rewards at every evaluated budget, with
the largest gains at small $N$, supporting budget mixtures as a way to
improve low-budget quality while retaining strong large-budget performance.

\begin{table}[t]
    \centering
    \small
    \caption{Best-of-$N$ rewards on GenEval2 and GenEval scores for the base SD3.5-M model,
    single Best-of-16 optimization, QUPO, FlowGRPO, and GRPO Guard. $\dagger$: FlowGRPO and GRPO Guard evaluation uses earlier checkpoints that match QUPO in GenEval score.}
    \label{tab:single_budget}
    \begin{tabular}{lccccccccc}
        \toprule
        Method & GenEval & Bo$1$ & Bo$4$ & Bo$8$ & Bo$16$ & Bo$32$ & Bo$64$ & Bo$96$ & Bo$128$ \\
        \midrule
        SD3.5-M     & 0.63 & 0.1804 & 0.3365 & 0.4193 & 0.4978 & 0.5679 & 0.6295 & 0.6625 & 0.6850 \\
        Single Bo16 & 0.75 & 0.1908 & 0.3518 & 0.4353 & 0.5131 & 0.5819 & 0.6417 & 0.6733 & 0.6946 \\
        QUPO        & \textbf{0.88} & \textbf{0.2287} & \textbf{0.3841} & \textbf{0.4618} & \textbf{0.5333} & \textbf{0.5962} & \textbf{0.6508} & \textbf{0.6794} & \textbf{0.7000} \\
        FlowGRPO$^\dagger$   & \textbf{0.88} & 0.2039 & 0.3256 & 0.3897 & 0.4550 & 0.5187 & 0.5783 & 0.6106 & 0.6322 \\
        GRPO Guard$^\dagger$ & \textbf{0.88} & 0.2013 & 0.3150 & 0.3740 & 0.4329 & 0.4901 & 0.5456 & 0.5773 & 0.5991 \\
        \bottomrule
    \end{tabular}
\end{table}

\subsection{Does Early Stopping Benefit RL Baselines?}

We next study whether early stopping improves the test-time scaling of RL baselines such as FlowGRPO and GRPO-Guard. We evaluate earlier checkpoints of both methods, selected to match QUPO's in-domain GenEval score of $0.88$. As shown in Table \ref{tab:single_budget}, early stopping does \textit{not} prevent frontier reversal: both baselines outperform SD3.5-M at Best-of-$1$ on GenEval2, but fall below it already at $N=4$. In contrast, QUPO maintains a positive estimated advantage through Best-of-$128$. These results show that weakened test-time scaling persists at matched in-domain performance and cannot be attributed simply to the baselines achieving higher GenEval scores. This supports our quantile-based interpretation: optimizing single-sample reward can improve typical generations without preserving the upper-tail quality exploited by Best-of-$N$ sampling.

\section{Diversity and Qualitative Results}
\label{app:additional}

\subsection{Perceptual Diversity}
\label{app:diversity}

We additionally examine whether weakened scaling transfer can be explained by a loss of perceptual diversity. Using the same $G=128$ GenEval2 generations per prompt for SD3.5-M and $G = 64$ for BAGEL models, we compute the average pairwise LPIPS distance~\cite{zhang2018unreasonable}
\begin{equation}
D(c)
=
\frac{2}{G(G-1)}
\sum_{1\leq i<j\leq G}
\mathrm{LPIPS}(\rvx_i,\rvx_j),
\end{equation}
and report $D(c)$ averaged across prompts. Higher values indicate greater within-prompt perceptual diversity.

\begin{table}[t]
\centering
\caption{\textbf{Within-prompt perceptual diversity on GenEval2.}
Average per-prompt pairwise LPIPS over all generated images.}
\label{tab:lpips_diversity}
\begin{tabular}{lc}
\toprule
Model & Mean pairwise LPIPS $\uparrow$ \\
\midrule
SD3.5-M & \textbf{0.6977} \\
FlowGRPO (official) & 0.5763 \\
QUPO & 0.6451 \\
\midrule
BAGEL & \textbf{0.6307} \\
UniGRPO & 0.5481 \\
QUPO & 0.5825 \\
\bottomrule
\end{tabular}
\end{table}

FlowGRPO substantially reduces perceptual diversity relative to the base model, while QUPO preserves considerably more of it. However, diversity alone does not explain Best-of-$N$ scaling: QUPO remains less diverse than the base model despite achieving a stronger large-$N$ reward frontier. Thus, diversity collapse can accompany upper-tail degradation, but perceptual diversity is not the quantity that directly determines scaling transfer. Our quantile analysis provides the more precise explanation: large-$N$ performance depends specifically on preserving rare {high-reward} outcomes, rather than diversity for its own sake.

\subsection{Additional Qualitative Comparisons}

\begin{figure}[!ht]
    \centering
    \begin{subfigure}[t]{0.49\linewidth}
        \centering
        \includegraphics[width=\linewidth]{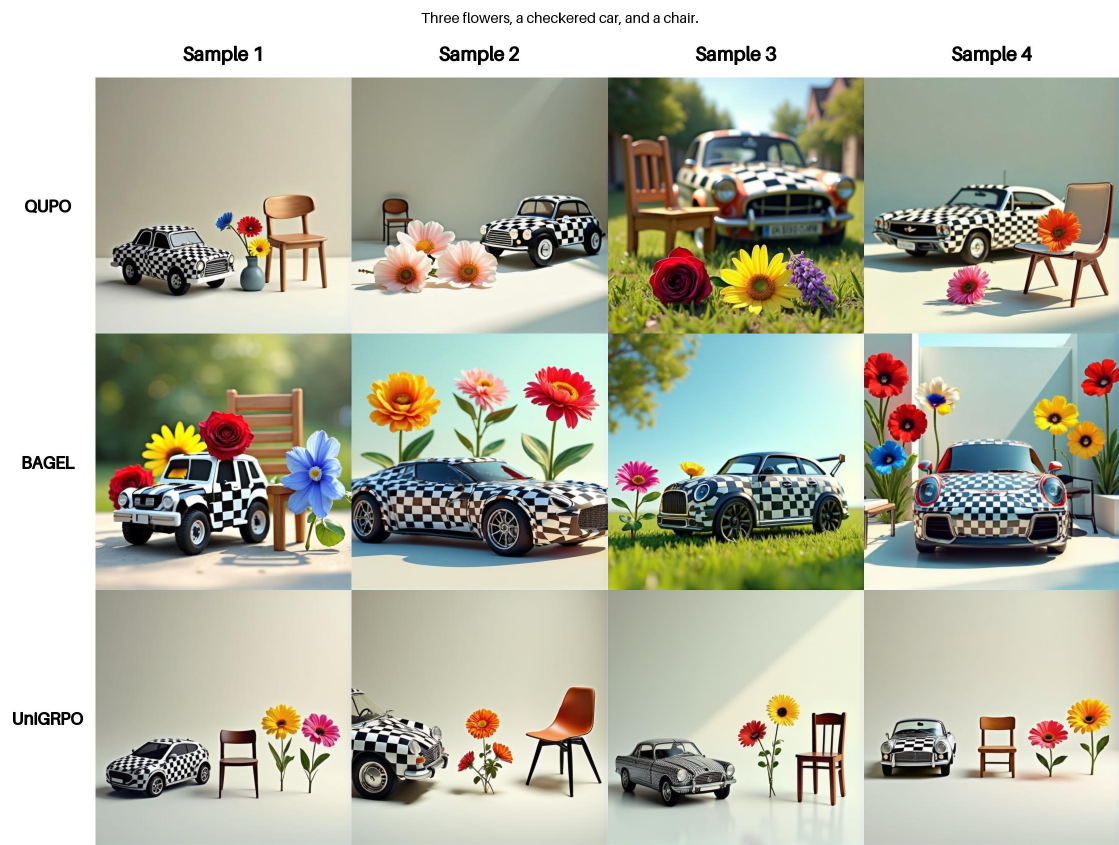}
        \caption{BAGEL on GenEval2.}
        \label{fig:bagel_qualitative}
    \end{subfigure}
    \hfill
    \begin{subfigure}[t]{0.49\linewidth}
        \centering
        \includegraphics[width=\linewidth]{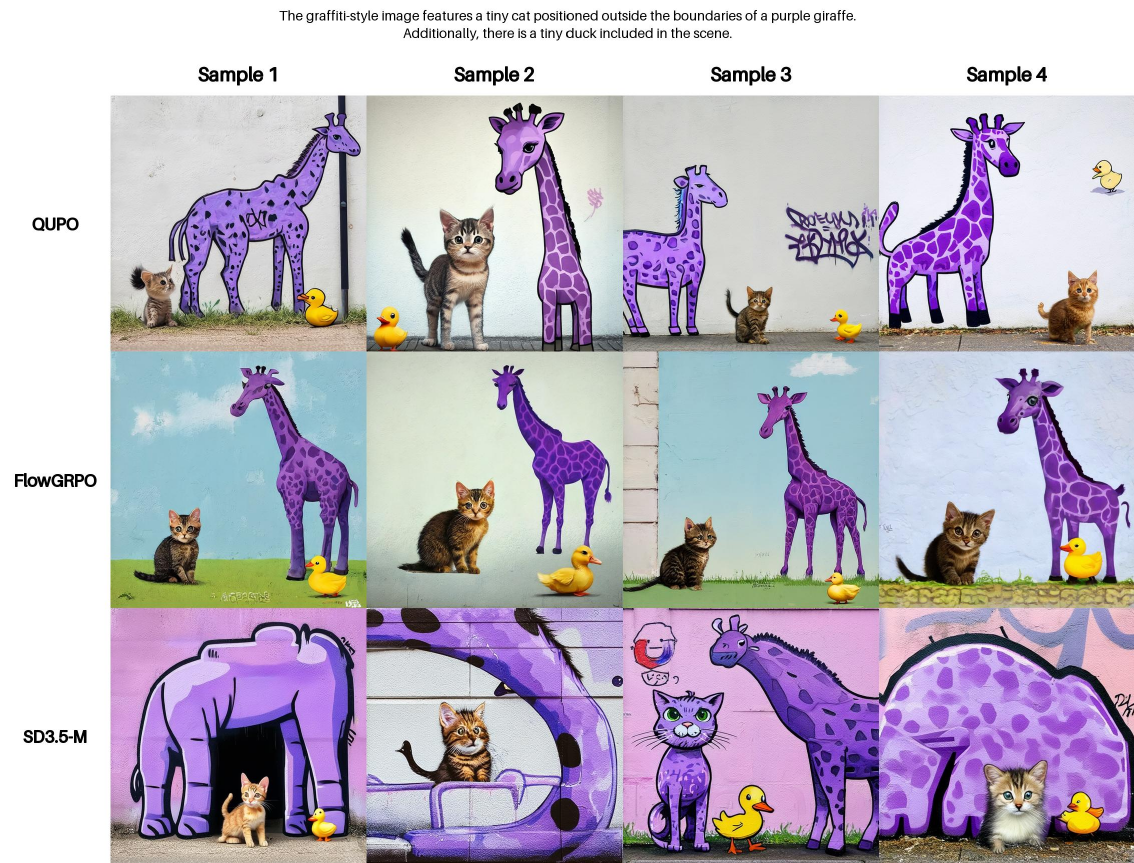}
        \caption{SD3.5-M on ConceptMix.}
        \label{fig:sd35_qualitative}
    \end{subfigure}

    \caption{\textbf{Qualitative examples of improved scaling transfer with QUPO.}
    \textbf{Left:} Four matched-seed generations from QUPO, base BAGEL, and UniGRPO. QUPO more consistently satisfies the requested composition, reflecting its strong scaling transfer.
    \textbf{Right:} Four matched-seed generations from QUPO, base SD3.5-M, and FlowGRPO on a held-out ConceptMix prompt. QUPO better preserves the requested composition and style; FlowGRPO often captures the three animals but misses the requested graffiti style.}
    \label{fig:qupo_qualitative}
\end{figure}

\begin{figure}[!ht]
    \centering
    \begin{subfigure}[t]{0.49\linewidth}
        \centering
        \includegraphics[width=\linewidth]{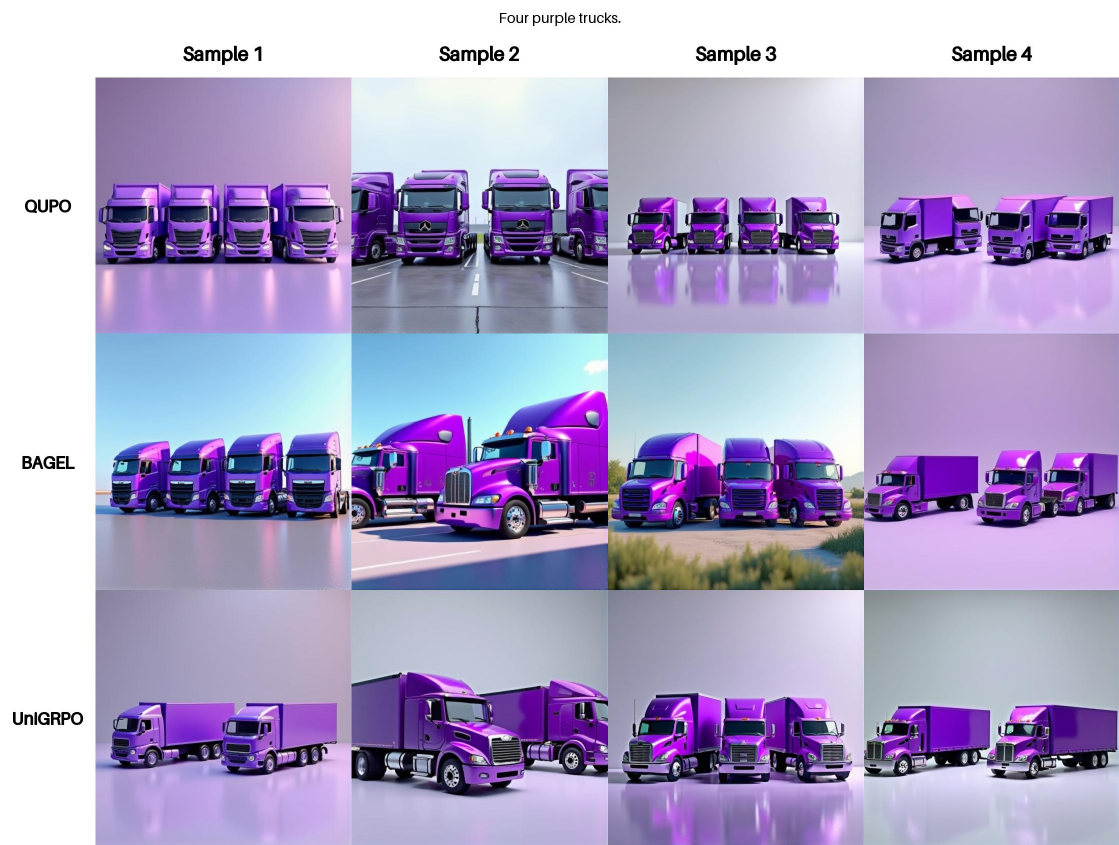}
        \caption{BAGEL on GenEval2.}
        \label{fig:bagel_qualitative}
    \end{subfigure}
    \hfill
    \begin{subfigure}[t]{0.49\linewidth}
        \centering
        \includegraphics[width=\linewidth]{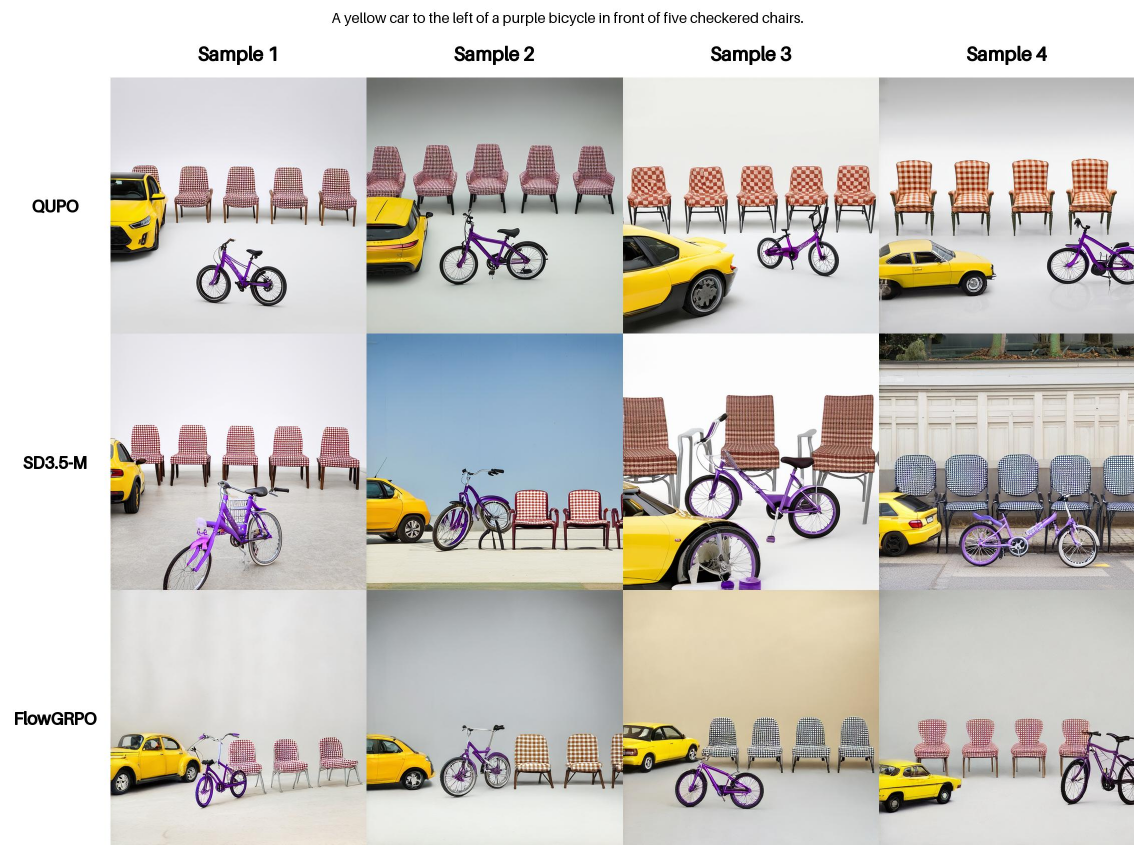}
        \caption{SD3.5-M on GenEval2.}
        \label{fig:sd35_qualitative}
    \end{subfigure}

    \caption{\textbf{Qualitative examples of improved scaling transfer with QUPO.}
    \textbf{Left:} Four matched-seed generations from QUPO, base BAGEL, and UniGRPO. QUPO more consistently satisfies the requested composition, reflecting its stronger scaling transfer.
    \textbf{Right:} Four matched-seed generations from QUPO, base SD3.5-M, and FlowGRPO on a held-out GenEval2 prompt.}
    \label{fig:qupo_qualitative_2}
\end{figure}

\begin{figure}[!ht]
    \centering
    \begin{subfigure}[t]{0.49\linewidth}
        \centering
        \includegraphics[width=\linewidth]{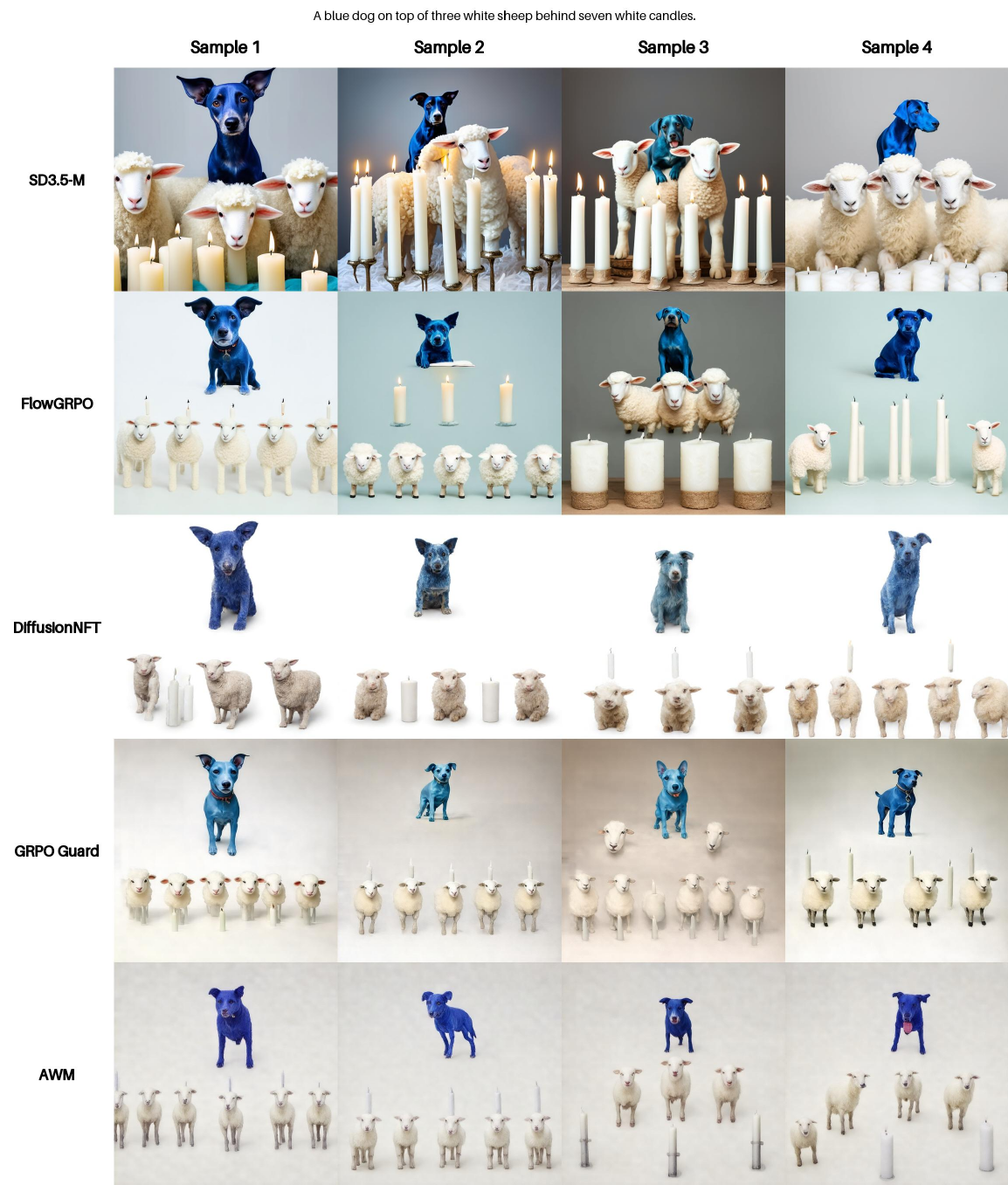}
        \caption{SD3.5-M GenEval2 comparison.}
        \label{fig:diagnostic_geneval2_qualitative_1}
    \end{subfigure}
    \hfill
    \begin{subfigure}[t]{0.49\linewidth}
        \centering
        \includegraphics[width=\linewidth]{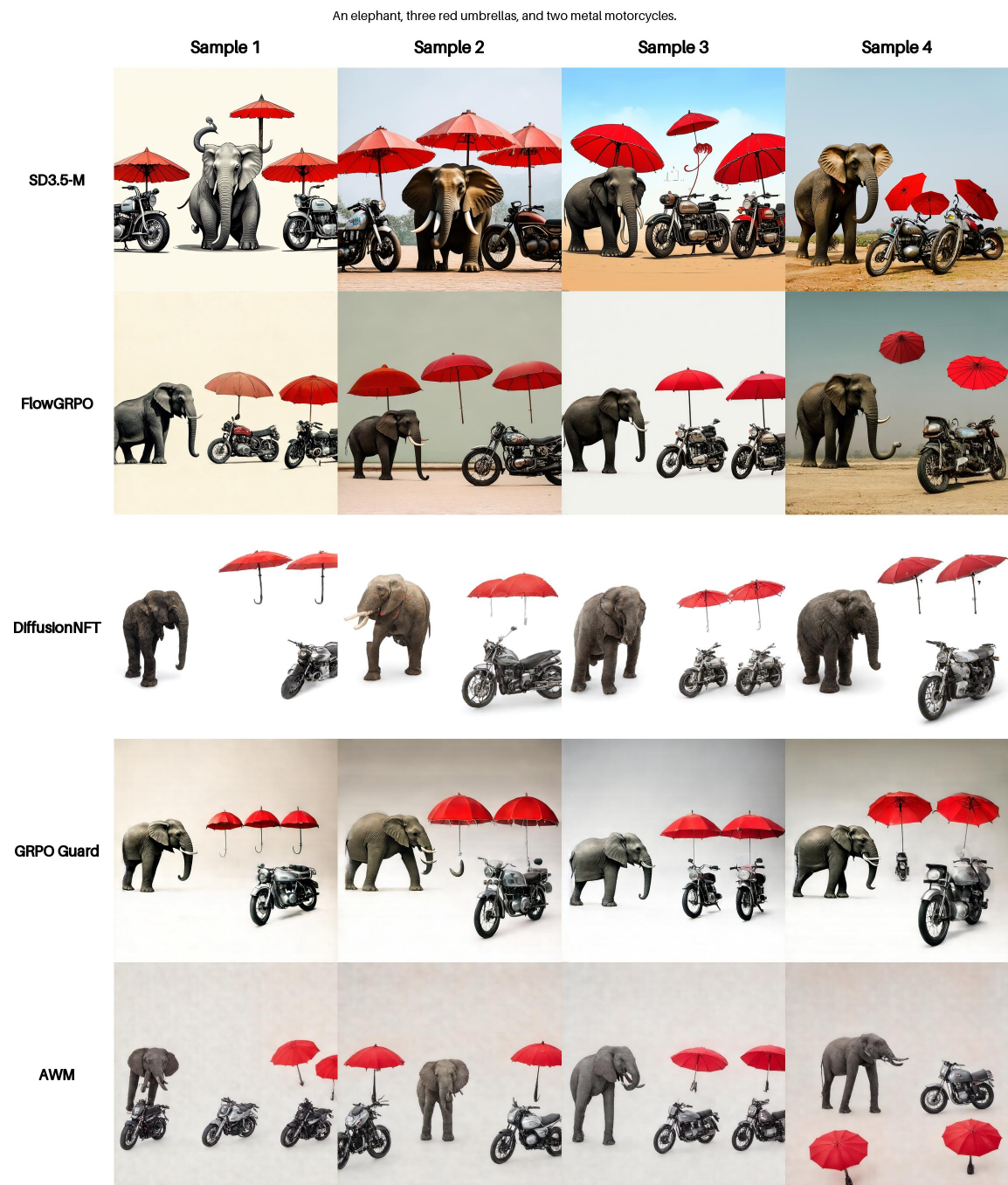}
        \caption{Additional SD3.5-M GenEval2 comparison.}
        \label{fig:diagnostic_geneval2_qualitative_2}
    \end{subfigure}

    \caption{\textbf{Multiple generations for the same held-out GenEval2 prompts.}
    Each row shows four independent samples from the base SD3.5-M model, FlowGRPO, DiffusionNFT, GRPO Guard, and AWM under matched generation seeds. In both examples, the base SD3.5-M model produces the rarer correct generations for these difficult prompts, illustrating the upper-tail failures that contribute to weakened Best-of-$N$ scaling after standard RL post-training.}
    \label{fig:diagnostic_geneval2_qualitative}
\end{figure}

We provide further qualitative examples from each model, across both the GenEval2 and ConceptMix benchmarks. As shown in Figures \ref{fig:qupo_qualitative} and \ref{fig:qupo_qualitative_2}, QUPO is able to more consistently generate the rare high-reward sample that benefits test-time scaling compared to its FlowGRPO and UniGRPO baselines. Figure \ref{fig:diagnostic_geneval2_qualitative} compares results of the base SD3.5-M model with the expected-reward RL baselines. As seen, SD3.5-M generates images that can capture the rare high-reward results more consistently, resulting in stronger scaling performance.  

\end{document}